\documentclass[a4paper,fleqn]{cas-sc}

\usepackage[authoryear]{natbib}
\usepackage{amsthm,amsmath,amssymb}
\usepackage{bbm}

\usepackage[ruled,vlined,linesnumbered]{algorithm2e}
\SetArgSty{textrm} 
\SetKwComment{Comment}{$\triangleright$ }{}

\usepackage{graphicx}
\usepackage{float}
\usepackage{subfigure}
\usepackage{caption}
\usepackage{threeparttable}
\usepackage{booktabs}
\usepackage{multirow}
\usepackage{array}
\usepackage{enumitem}
\usepackage{soul, color, xcolor}

\usepackage{utfsym}

\makeatletter
\renewcommand{\fnum@figure}{Fig. \thefigure.\@gobble }
\makeatother

\def\tsc#1{\csdef{#1}{\textsc{\lowercase{#1}}\xspace}}
\tsc{WGM}
\tsc{QE}
\tsc{EP}
\tsc{PMS}
\tsc{BEC}
\tsc{DE}

\begin{document}
\let\WriteBookmarks\relax
\def\floatpagepagefraction{1}
\def\textpagefraction{.001}
\let\printorcid\relax

\shorttitle{}

\shortauthors{Xinzheng Wu et~al.}

\title [mode = title]{VLM as Strategist: Adaptive Generation of Safety-critical Testing Scenarios via Guided Diffusion}                    



%
\author[1]{\textcolor{black}{Xinzheng Wu}}[]
\credit{Formal analysis, Methodology, Visualization, Writing - original draft}
\affiliation[1]{organization={School of Automotive Studies, Tongji University},
    addressline={No. 4800 Cao'an Road.}, 
    city={Shanghai},
    postcode={201804}, 
    country={China}}

\author[1]{\textcolor{black}{Junyi Chen}}[]
\credit{Formal analysis, Project administration, Validation, Writing - review \& editing}
\cormark[1]
\ead{chenjunyi@tongji.edu.cn}

\author[1]{\textcolor{black}{Naiting Zhong}}[]
\credit{Data curation, Software, Validation, Writing - review \& editing}

\author[1]{\textcolor{black}{Yong Shen}}[]
\credit{Investigation, Resources, Supervision}


\cortext[cor1]{Corresponding author}

\begin{abstract}
The safe deployment of autonomous driving systems (ADSs) relies on comprehensive testing and evaluation. However, safety-critical scenarios that can effectively expose system vulnerabilities are extremely sparse in the real world. Existing scenario generation methods face challenges in efficiently constructing long-tail scenarios that ensure fidelity, criticality, and interactivity, while particularly lacking real-time dynamic response capabilities to the vehicle under test (VUT).
To address these challenges, this paper proposes a safety-critical testing scenario generation framework that integrates the high-level semantic understanding capabilities of Vision Language Models (VLMs) with the fine-grained generation capabilities of adaptive guided diffusion models. The framework establishes a three-layer hierarchical architecture comprising a strategic layer for VLM-directed scenario generation objective determination, a tactical layer for guidance function formulation, and an operational layer for guided diffusion execution. We first establish a high-quality fundamental diffusion model that learns the data distribution of real driving scenarios. Next, we design an adaptive guided diffusion method that enables real-time, precise control of background vehicles (BVs) in closed-loop simulation. The VLM is then incorporated to autonomously generate scenario generation objectives and guidance functions through deep scenario understanding and risk reasoning, ultimately guiding the diffusion model to achieve VLM-directed scenario generation. Experimental results demonstrate that the proposed method can efficiently generate realistic, diverse, and highly interactive safety-critical testing scenarios. Compared with original scenarios, the generated critical scenarios increase the average at-fault collision rate of the AUTs by approximately $4.2\times$. Furthermore, case studies validate the adaptability and VLM-directed generation performance of the proposed method.
\end{abstract}




\maketitle

\section{Introduction} \label{sec1}
Autonomous driving technology is spearheading a transformation in the global automotive industries, and its safe and reliable implementation is the core prerequisite for large-scale adoption \citep{ren2025intelligent}. Comprehensive testing and evaluation of autonomous driving systems (ADSs) are essential to ensuring their safety, in which the identification and generation of safety-critical scenarios represent a core challenge \citep{yang2025adaptive}. "Safety-critical scenarios" specifically refer to rare driving situations with potentially high risks \citep{ding2023surveya}. Conducting tests under such scenarios enables effective evaluation of the ADSs' safety performance, as well as the clarification and iterative refinement of its Operational Design Domain (ODD). However, due to the rarity of safety-critical scenarios in naturalistic driving environments \citep{feng2023dense}, real-world road testing is inefficient and cost-prohibitive, making it unsuitable for large-scale testing of high-level ADSs.

As a more efficient and practical solution, simulation-based testing has garnered significant industrial and scholarly attention \citep{sun2022scenariobased}. In recent years, engineers in enterprises generally extract safety-critical testing scenarios by directly replaying vehicle-collected data in simulation environments \citep{liu2024survey}, while some researchers achieve accelerated sampling of safety-critical scenarios through optimization-based search within a predefined scenario parameter space \citep{wu2024accelerated, wu2026make}. However, the background vehicles (BVs) in the safety-critical testing scenarios generated by the aforementioned methods exhibit fixed behaviors and cannot dynamically respond to the actions of the vehicle under test (VUT). As a remedy, some other studies have introduced reinforcement learning to train adversarial BV driver models, thereby constructing naturalistic adversarial driving environments (NADE) \citep{feng2021intelligenta} or evolving scenarios \citep{ma2024evolving, wu2025evolving}. These methods generate safety-critical testing scenarios through adversarial interactions between BVs and the VUT, but they often struggle to achieve a balance between scenario adversariality and realism.

More recently, deep generative models, particularly diffusion models, have become a research hotspot in the field of scenario generation and closed-loop simulation due to their exceptional capability in learning complex data distributions \citep{yang2024drivearena, mei2025dreamforge}. Moreover, numerous studies have achieved controllable generation of driving trajectories based on guided diffusion models while ensuring realism \citep{zhong2022guided, jiang2023motiondiffuser}. Further, large models (LMs), including large language models (LLMs) and vision language models (VLMs), can also be incorporated to enable more flexible and user-friendly scenario generation \citep{zhong2023languageguided, peng2025ldscene}. The aforementioned research demonstrates the significant promise of LM-guided diffusion models for scenario generation. However, existing studies predominantly employ static diffusion guidance, where guidance strategies are fixed at the beginning of simulation, preventing BVs' behavioral policies from dynamically adjusting to the VUT's real-time actions during simulation. Concurrently, existing research confines the application of LMs merely to "translators", where LMs only convert user-defined vehicle behavior instructions into underlying guidance functions. This approach is fundamentally a \textit{user-specified} scenario generation paradigm, in which the types of generated scenarios remain constrained by expert experience, overlooking LMs' formidable capabilities in scenario understanding and reasoning. These limitations fail to meet the requirements for real-time interaction and behavioral diversity in safety-critical testing scenarios.

\begin{figure}[pos=b] 
      \centering
      \includegraphics[width=16cm]{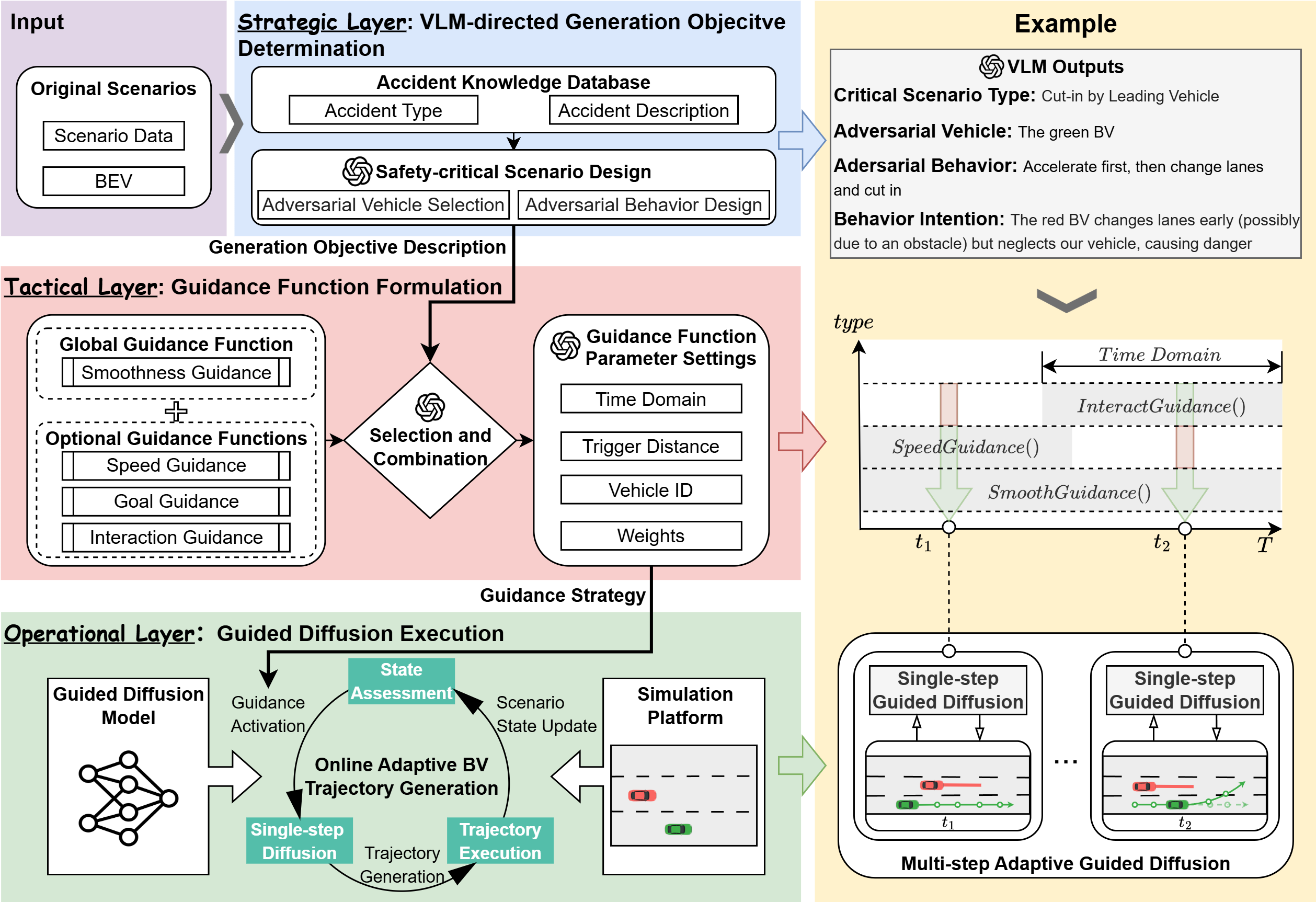}
      \caption{Framework of the proposed method.}
      \label{Framework}
\end{figure} 

To address the aforementioned challenges, this paper proposes an \textit{VLM-directed} adaptive generation method based on guided diffusion model. As shown in Fig. \ref{Framework}, a three-layer hierarchical architecture is proposed for safety-critical testing scenario generation. At the strategic layer, unlike the aforementioned \textit{user-specified} approaches that treat LMs merely as translators, the \textit{VLM-directed} method proposed in this paper positions LMs as strategists. Users only need to specify the top-level direction for scenario generation (e.g., generating collision scenarios), and LMs can leverage prior knowledge to accomplish scenario understanding, adversarial vehicle selection, and adversarial intention and behavior reasoning. It is particularly noteworthy that we use both descriptive information of the scenario and the Bird's-Eye Views (BEVs) at the start and end moments of the scenario as input, and employ VLMs to achieve a more comprehensive understanding of the original scenario. 

At the tactical layer, to avoid the low usability and high unpredictability associated with LLMs directly generating code, this study requires VLMs to select and combine predefined guidance function templates based on the scenario generation objectives from the strategic layer, and to further configure the parameters of the guidance functions. Specially, to achieve adaptive guidance for BVs in safety-critical testing scenarios, we introduce the concept of a time window into the guidance functions. This involves configuring the time domain and trigger distance of the guidance functions, enabling the behavior of BVs to dynamically respond to the actions of the VUT.

Finally, at the operational layer, a fundamental guided diffusion model is trained. Based on the guidance function defined at the tactical layer, the trajectories of the BVs are generated and executed after a single-step guided diffusion process. The above procedure is iteratively carried out in the simulation platform, ultimately achieving multi-step adaptive guided diffusion and closed-loop simulation testing. 
The main contributions of this study can be summarized as follows:

\begin{enumerate}[label=(\arabic*)]
  \item We propose a strategic-tactical-operational scenario generation framework. By positioning VLMs as strategists instead of translators, this framework can fully leverage the scenario understanding and reasoning capabilities of VLMs to achieve VLM-directed generation of diverse safety-critical testing scenarios, while effectively circumventing the inherent subjectivity and limitations of the user-specified methods.
  \item We develop an adaptive guided diffusion method. This method adaptively adjusts the guidance strategy by defining the time domain and trigger distance for different guidance functions based on the scenario state. Meanwhile, during simulation, it enables online updates of BV trajectories through a multi-step guided diffusion process. By doing so, the BV can respond to the SV's behavior in real time and continuously apply adversarial interactions.
  \item We construct a co-simulation platform supporting closed-loop testing and conduct extensive experiments based on the nuPlan dataset. The experiments demonstrate that the proposed method significantly enhances the criticality of original scenarios, while also validating its adaptability to diverse scenarios and different VUTs controlled by four algorithms under test (AUTs), as well as its performance in VLM-directed scenario generation. 
\end{enumerate}

The remainder of this paper is structured as follows: Section \ref{sec2} introduces and analyzes the related works. In Section \ref{sec3}, the construction of the three-layer hierarchical framework is described in detail. Section \ref{sec4} puts the proposed method into application with four different VUTs and evaluate its generation performance, while ablation experiment on VLMs is conducted. Finally, Section \ref{sec5} concludes this paper.

\section{Related works} \label{sec2}

\subsection{Generation of safety-critical testing scenarios for ADSs}

In the current literature, scenarios for testing ADSs primarily challenge the VUT by controlling the behavior of BVs. Based on the methodology for determining BV behavior, the generation of safety-critical testing scenarios for ADSs can be categorized into non-interactive methods and interactive methods. In non-interactive methods, combinatorial testing is a typical representative. This method achieves comprehensive coverage of safety-critical scenarios in the parameter space through systematic combination of scenario parameters, thereby enabling safety validation of ADSs \citep{li2023simulationbased, zhou2023flyover, diemert2023challenging}. However, for the high-dimensional parameter space of ADS scenarios, achieving exhaustive testing requires substantial computational resources. As a remedy, many researchers have proposed accelerated testing approaches that leverage optimized sampling to rapidly search for critical scenarios within the scenario parameter space \citep{gong2023adaptivea, wu2024accelerated, wu2026make}. Nevertheless, due to the inherent limitations of non-interactive methods, the behavior of BVs cannot respond in real-time to the actual actions of the VUT, which often leads to reduced usability of test cases.

Interactive methods address the aforementioned issues by modeling BVs as driver models with independent decision-making and planning (D\&P) capabilities. In early research, rule-based models such as IDM \citep{treiber2000congested}, Nilsson \citep{nilsson2016if}, and Stackelberg \citep{li2022gametheoretic} are widely used in traffic simulation. However, these models lack the adversarial nature required to generate safety-critical scenarios. In recent years, reinforcement learning-based methods have become increasingly popular. Researchers train adversarial BV driver models to create adversarial testing environments \citep{feng2021intelligenta, feng2023dense, ren2025intelligent} or evolving scenarios \citep{ma2024evolving, wu2025evolving}, enabling BVs to autonomously generate safety-critical scenarios through continuous interaction with the VUT. However, reinforcement learning-based methods require high training costs, and how to demonstrate the behavioral fidelity of the trained driver models remains an open question. In response to this challenge, a growing number of researchers are adopting diffusion models to generate scenarios.

\subsection{Diffusion models for scenario generation}
Recent advancements in diffusion models provide new solutions to generate scenarios with both high fidelity and high interaction \citep{peng2025diffusion}. Due to the ability to effectively capture the intrinsic multi-modality of data distributions, diffusion models can more comprehensively learn the underlying distribution of scenarios, making them highly promising for generating realistic scenarios \citep{chen2024datadriven}. Furthermore, many researchers are dedicated to developing controllable diffusion models \citep{jiang2024survey} to achieve guided generation of driving trajectories, thereby achieving interactive scenarios generation. For instance, MotionDiffuser \citep{jiang2023motiondiffuser} proposes a general constrained sampling framework that enables controlled trajectory sampling based on differentiable cost functions. CTG \citep{zhong2022guided} uses STL (Signal Temporal Logic) and defined rules (e.g., reach a goal or follow a speed limit) as objective functions to guide the diffusion model.

The aforementioned research lays the foundation for generating both realistic and safety-critical testing scenarios: since trajectory generation is controllable, simply setting the guidance objective as safety-critical enables the control of BVs to challenge the VUT. Following this idea, DiffScene \citep{xu2025diffscene} employs three guidance objectives to maximize the driving risk of the VUT while controlling BVs to prevent the VUT from completing its driving tasks. Similarly, Lu et al. \citep{lu2024datadriven} uses guidance objectives to control the behavior complexity of BVs and the traffic density. Moreover, VBD \citep{huang2024versatile} utilizes conflict-prior guidance and game-theoretic guidance respectively to facilitate the generation of safety-critical scenarios.
However, existing research still primarily focuses on the controllability of the generation process, while studies on what kinds of safety-critical scenarios should be generated for effective testing and the diversity of generated safety-critical scenarios remain insufficient.

\subsection{Application of large models in ADSs testing}
The emerging large model technology, represented by LLMs and VLMs, has injected new vitality into testing scenario generation due to their powerful commonsense reasoning and multi-modal understanding capabilities \citep{yang2024llm4drive, cui2024survey}. As an intuitive way, some studies directly employ LLMs to translate natural language instructions into scenario parameters or scripts \citep{zhao2024chat2scenario, zhang2024chatscene}, or to extract risk factors from accident reports for generating high-risk scenarios \citep{lu2024realistic}. However, the primary strength of LMs lies in high-level scenario comprehension and reasoning, while they are less adept at fine-grained generation of low-level scenario parameters or trajectories.

To fill this gap, other studies utilize LLMs for high-level decision-making to guide the aforementioned underlying models in generating adversarial or specific scenarios. For example, LLM-Attacker \citep{mei2025llmattacker} employs LLMs as adversarial BV identifiers and integrates a pre-trained probabilistic traffic forecasting model to generate trajectories for adversarial BVs. Moreover, CTG++ \citep{zhong2023languageguided} and LD-scene \citep{peng2025ldscene} harness LLMs to convert users' queries into the codes of loss functions, which then guide the diffusion model toward query-compliant generation. However, their application in generating safety-critical testing scenarios remains \textit{behavior-oriented}, where both the types of safety-critical scenarios and the adversarial behaviors of BVs require manual specification by users. This reliance on expert knowledge limits the diversity of generated safety-critical scenarios. Moreover, these methods rely solely on LLMs, exhibiting inherent limitations in handling complex visual scenario elements.

To overcome the visual comprehension limitations of LLMs, researchers start to explore the application of VLMs. For instance, DriveGen \citep{zhang2025drivegen} employs VLM to interpret BEV representations of scenarios. Based on its understanding of scenario layouts and vehicle positioning, it establishes appropriate target destinations for BVs, thereby guiding diffusion models to generate corresponding trajectories. Gen-Drive \citep{huang2025gendrive} incorporates VLMs to evaluate the plausibility of generated scenarios, then feeds the evaluation results back to the generative model to guide the generation of scenarios. The aforementioned approaches demonstrate VLMs' powerful scenario comprehension and reasoning capabilities, which show promising potential for achieving VLM-directed safety-critical testing scenario generation.

In summary, based on the aforementioned discussion, this paper employs a VLM to determine high-level safety-critical scenario generation objectives, which subsequently guides the underlying diffusion model to achieve controllable fine-grained trajectory generation and finally enabling VLM-directed adaptive generation of safety-critical testing scenarios.

\section{Methodology} \label{sec3}

In this section, we present a three-layer hierarchical framework for safety-critical testing scenario generation, as illustrated in Fig \ref{Framework}. To facilitate better understanding, we adopt a bottom-up way to elaborate the framework: We first introduce the construction of the fundamental diffusion model for scenario generation. Then, we detail the design of the adaptive guidance framework and the guidance function templates. Finally, we describe the VLM-based determination of scenario generation objectives and the formulation of guidance functions.

\subsection{Problem formulation}

This paper aims to generate safety-critical testing scenarios for ADSs while ensuring scenario fidelity, diversity, and interactivity. Consider a scenario from a certain dataset $\mathbbm{D}$ with time horizon $T$. It includes one subject vehicle (SV, which will be replaced by the AUT-controlled VUT during the generation), $N$ traffic participants (hereinafter referred to as agents), a map $\boldsymbol{m}$ and traffic signals $Tl$. 
The state sequence of all agents is denoted as $S=[s_{0:T}^1, ..., s_{0:T}^N]$, with $s_t^i =[type^i, \boldsymbol{pos}^i,h_t^i, \boldsymbol{v}_t^i, \boldsymbol{bbox}^i]$ (type, position, heading, velocity, 3d size) representing the state of the $i$-th agent at timestep $t$. And the action sequence of all agents is denoted as $A = [a_{0:T}^1, ..., a_{0:T}^N]$, where $a_t^i = [\dot{\boldsymbol{v}}_t^i, \dot{\theta}^i_t]$ (acceleration, yaw rate) represents the action of the $i$-th agent at timestep $t$. 

Unlike D\&P tasks that control the SV, the testing scenario generation task achieves testing objectives by controlling one or more BVs to challenge the VUT. In this paper, adversarial BVs are selected by the VLM based on scenario understanding, and their future trajectories will satisfy the guidance function $\mathcal{G}$. 
Furthermore, to ensure the interactivity of testing scenarios, this paper adopts an online closed-loop simulation paradigm. Specifically, during scenario generation, the future trajectories for adversarial agents are updated and executed at a frequency $f_{\text{rep}}$. 

More concretely, at the simulation timestep $t_{sim}$ where replanning is required, given $\boldsymbol{m}$, $Tl$, and historical trajectories of all agents $S_{t_{sim}-T_{hist}:t_{sim}-1}$ as scenario context $\boldsymbol{c}$, our model employs the learned scenario distribution $P(A_{t_{sim}+1:t_{sim}+T_{fut} } \vert \boldsymbol{c})$ to predict future trajectories $S_{t_{sim}+1:t_{sim}+T_{fut}}=f(A_{t_{sim}+1:t_{sim}+T_{fut}})$ for all agents, while ensuring that the future trajectories of VLM-selected adversarial BVs comply with the guidance function $\mathcal{G}$. Here, $T_{hist}$ denotes the time horizon of the historical trajectories required as model input, $T_{fut}$ represents the time horizon of the future trajectories output by the model, and $f$ represents the dynamics model.

\subsection{Fundamental diffusion model for scenario generation}
\subsubsection{Denoising diffusion probabilistic model}
Under the basic framework of diffusion models, Denoising Diffusion Probabilistic Model (DDPM) \citep{ho2020denoising} is one of the most influential implementations. It consists of two key processes: a fixed forward diffusion process and a learnable reverse denoising process. In the forward process, Gaussian noise is gradually injected into the original action data $A^{(0)}$ over $K$ diffusion steps. Note that while $A_t$ represents the actions of all traffic participants at a specific timestep, $A^{(k)}$ denotes the noised action sequence of all participants after the $k$-th diffusion step. Correspondingly, $A^{(0)}$ is the original noise-free action sequence, and $A^{(K)}$ represents the final diffused action sequence that approximates an isotropic Gaussian distribution. The aforementioned forward process can be described as:


\begin{align}
    \label{eq1}
    q(A^{(1:K)}\vert A^{(0)}) & := \prod_{k=1}^K q(A^{(k)} \vert A^{(k-1)}) \\
    \label{eq2}
    q(A^{(k)} \vert A^{(k-1)}) &  := \mathcal{N}\left(A^{(k)}; \sqrt{1-\beta_k} A^{(k-1)}, \beta_k \textbf{I}\right)
\end{align}

where $\beta_k \in (0,1)$ is a predefined noise schedule that controls the noise level at each diffusion step. By expanding Eq. \ref{eq1} and substituting Eq. \ref{eq2} into it, the following simplified expression can be derived:

\begin{equation}
\label{eq3}
\begin{aligned}
    A^{(k)} = \sqrt{\bar{\alpha}_k} A^{(0)} + \sqrt{1-\bar{\alpha}_k} \epsilon
\end{aligned}
\end{equation}
where $\alpha_k = 1-\beta_k$, $\bar{\alpha}_k=\prod_{i=1}^k\alpha_i$, and $\epsilon \sim \mathcal{N}(0, \textbf{I})$ follows a multivariate Gaussian distribution with zero mean and identity covariance matrix $\textbf{I}$. 
According to Eq. \ref{eq3}, we can directly obtain the noised action sequence $A^{(k)}$ at any diffusion step from the original $A^{(0)}$. Additionally, for the noise schedule $\beta_k$, common choices include linear \citep{ho2020denoising}, cosine \citep{nichol2021improved}, and log \citep{huang2024versatile} schedules. In this work, we adopt the cosine schedule.

In the reverse process, DDPM learns to iteratively denoise from a Gaussian noise $\tilde{A}^{(K)} \sim \mathcal{N}(0,\textbf{I})$  until recovering a realistic action sequence $\tilde{A}^{(0)}$ that matches the true data distribution. Note that we use $\tilde{A}^{(k)}$ to denote the action sequence obtained during the reverse denoising process. The aforementioned process can be mathematically represented as:

\begin{align}
    \label{eq4}
    & p_{\theta}(\tilde{A}^{(0:K)}\vert \boldsymbol{c})  := q(\tilde{A}^{(K)})\prod_{k=1}^K p_\theta(\tilde{A}^{(k-1)} \vert \tilde{A}^{(k)}, \boldsymbol{c})  \\
    \label{eq5}
    & p_\theta(\tilde{A}^{(k-1)} \vert \tilde{A}^{(k)}, \boldsymbol{c}) : = \mathcal{N}(\tilde{A}^{(k-1)}; \mu_k, \sigma^2_k\textbf{I}) 
\end{align}

As shown in Eq. \ref{eq4}, the reverse denoising process proceeds iteratively, where the core of each step lies in estimating the mean $\mu_k$ and variance $\sigma^2_k$ of the Gaussian distribution from step $k$ to $k-1$ in Eq. \ref{eq5}. According to \cite{ho2020denoising}, these values can be computed as:

\begin{align}
    \label{eq6}
    & \mu_k =  \frac{\sqrt{\bar{\alpha}_{k-1}} \beta_k}{1 - \bar{\alpha}_k} \hat{A}^{(0)} + \frac{\sqrt{\alpha_k} (1 - \bar{\alpha}_{k-1})}{1 - \bar{\alpha}_k} \tilde{A}^{(K)} =  \frac{1}{\sqrt{\alpha_k}} \left( \tilde{A}^{(K)} - \frac{1 - \alpha_k}{\sqrt{1 - \bar{\alpha}_k}} \, \hat{\epsilon} \right)\\
    \label{eq7}
    & \sigma^2_k = \frac{1-\bar{\alpha}_{k-1}}{1-\bar{\alpha}_k} \beta_k
\end{align}

According to Eq. \ref{eq7}, the variance $\sigma^2$ can be directly computed based on the predefined noise schedule $\beta_k$ during the forward diffusion process. Therefore, only $\hat{A}^{(0)}$ or $\hat{\epsilon}$ in Eq. \ref{eq6} requires estimation. In the original DDPM framework, a neural network $\epsilon_\theta$ with parameters $\theta$ is trained to predict $\hat{\epsilon} = \epsilon_\theta(\tilde{A}^{(k)}, k, \boldsymbol{c})$, conditioned on the noised input $\tilde{A}^{(k)}$, the denoising step $k$, and the scenario context $\boldsymbol{c}$. While in this paper, similar to \cite{huang2024versatile}, we train a denoiser $\mathcal{D}_\theta$ to directly predict the noise-free action sequence $\hat{A}^{(0)} = \mathcal{D}_\theta(\tilde{A}^{(k)}, k, \boldsymbol{c})$ and then calculate $\mu_k$.

\subsubsection{Model structure}

\begin{figure}[pos=t] 
      \centering
      \includegraphics[width=16cm]{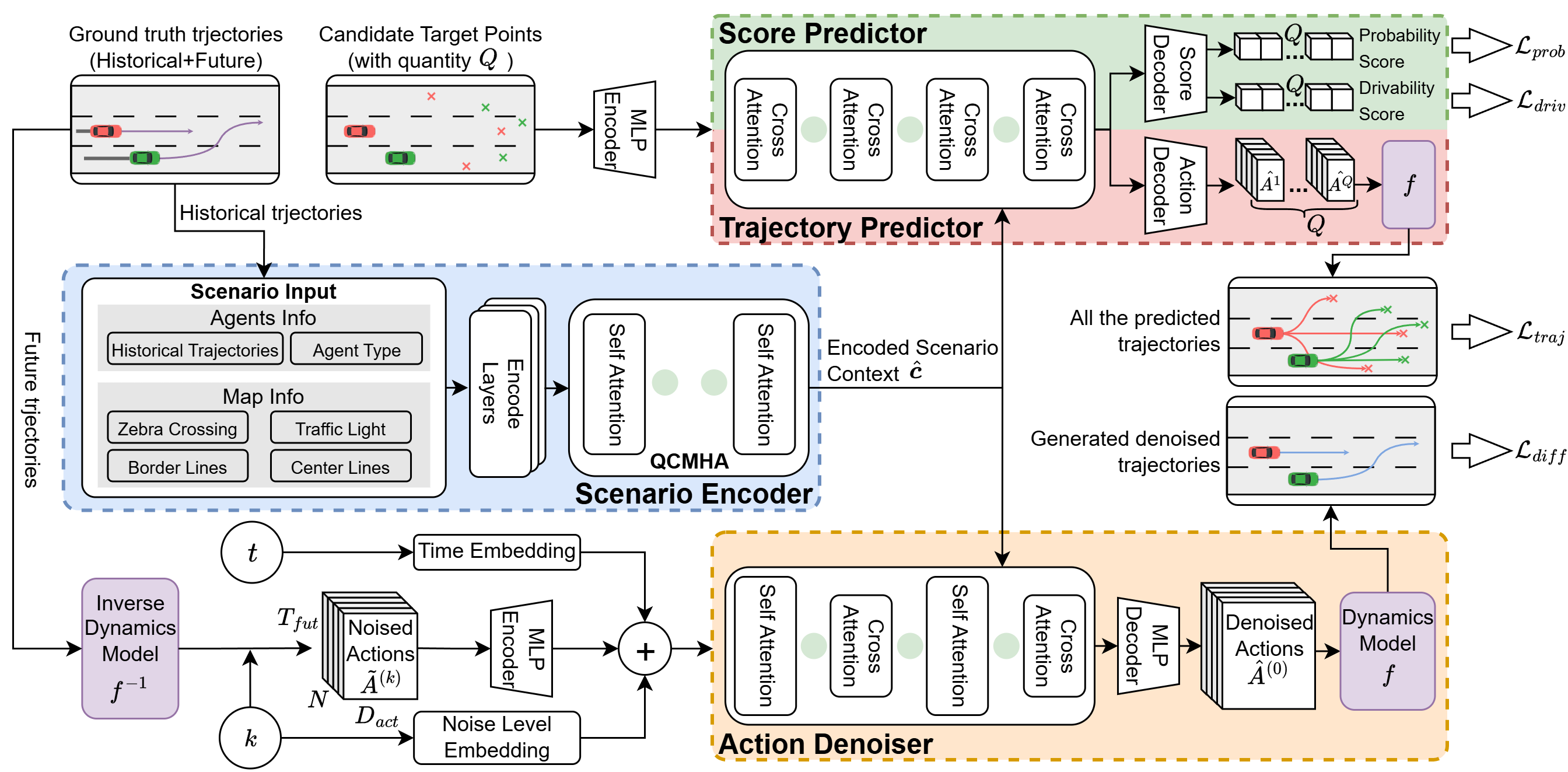}
      \caption{Structure of the fundamental diffusion model.}
      \label{Diff_structure}
\end{figure}

The structure of the fundamental diffusion model is illustrated in Fig. \ref{Diff_structure}. As shown in the figure, the proposed model adopts the prevailing encoder-decoder architecture consisting of two core modules: a Scenario Encoder and an Action Denoiser. Additionally, inspired by \cite{huang2024versatile}, this paper introduces auxiliary training objectives alongside the main generation task. Specifically, we design a Trajectory Predictor and a Score Predictor downstream of the Scenario Encoder. These auxiliary modules leverage the encoded scenario context $\hat{\boldsymbol{c}}$ to perform explicit trajectory prediction or scenario assessment. The losses generated by these modules are combined with the main diffusion model’s loss to jointly optimize the model, aiming to encourage the Scenario Encoder to learn more informative feature representations. 

\textbf{Scenario encoder}: The goal of the scenario encoder $\mathcal{E}_\theta$ is to encode dynamic agent historical trajectories and static map information (including lane markings and traffic signals) into a unified latent space feature $\hat{\boldsymbol{c}}$, providing comprehensive contextual conditioning for the subsequent diffusion model and auxiliary tasks. 

The scenario encoder first processes each scenario element through different encoding layers to generate initial representations $\hat{\boldsymbol{e}}$ with unified dimensionality. Specifically, for agent historical trajectories, combined with agent-type embeddings, a shared multilayer perceptron (MLP) network is utilized to extracts temporal dynamic features along the time dimension and outputs $\hat{\boldsymbol{e}}_{agent}$. For map lane markings, each point is first encoded using an MLP, followed by max-pooling along the point sequence dimension to obtain $\hat{\boldsymbol{e}}_{map}$. For traffic signal information, an MLP is also applied to generate $\hat{\boldsymbol{e}}_{tl}$. Furthermore, to explicitly capture relationships $\hat{\boldsymbol{r}}$ between scenario elements, the scenario encoder incorporates Fourier embeddings \citep{zhu2021fourier}, which map geometric relationship between agents and between agents and map elements into the high-dimensional latent space.

After obtaining the initial representations $\hat{\boldsymbol{e}}$ of all scenario elements and their relationships $\hat{\boldsymbol{r}}$, this information is fed into a multi-layer Transformer encoder network for further modeling. The network consists of multiple self-attention Transformer blocks. Each block incorporates a Query-Centric Multi-Head Attention (QCMHA) module \citep{huang2024dtpp}, along with standard components including feed-forward neural networks, layer normalization, and residual connections. Through this encoding pipeline, the scenario encoder ultimately outputs a unified scenario context tensor $\hat{\boldsymbol{c}}$.

\textbf{Action denoiser}: As the core component of the diffusion model, the action denoiser $\mathcal{D}_\theta$ is designed to predict the original noise-free actions $\hat{A}^{(0)}$ based on the noised action sequence $\tilde{A}^{(k)}$, the current noise level $k$, and the contextual features $\hat{\boldsymbol{c}}$ provided by the scenario encoder. 

As shown in Fig. \ref{Diff_structure}, during the training phase, the input to the action denoiser is derived from the ground-truth future trajectories. These trajectories are first processed by an inverse dynamics model $f^{-1}$ to compute the original action sequence $A^{(0)}=f^{-1}(S_{T_{hist}+1:T})$, which is then diffused through Eq. \ref{eq3} to generate the noised action sequence $\tilde{A}^{(k)}\in \mathbbm{R}^{N\times T_{fut} \times D_{act}}$ at a given noise level $k$, where $N$ is the number of agents, $T_{fut}$ represents the length of future trajectories, and $D_{act} = 2$ is the dimension of the action. After that, $\tilde{A}^{(k)}$ is encoded by an MLP encoder, and the resulting encoding is concatenated with time embeddings and noise level embeddings before being fed into the action denoiser. Furthermore, during the inference phase, the input to the action denoiser is directly a pure noisy action sequence $\tilde{A}^{(K)}$.

The internal structure of the action denoiser is primarily based on a Transformer network incorporating both self-attention and cross-attention blocks. The self-attention blocks are designed to model the joint distribution of all agents' future states and capture their mutual interactions, while employing causal masking \citep{huang2024dtpp} to ensure temporal causality in predictions. The cross-attention blocks are designed for correlating the noised trajectory information with the scenario context $\hat{\boldsymbol{c}}$, thereby modeling the action distribution under scenario constraints. By doing this, scenario information, such as map constraints and other agents' behaviors, can be used to effectively guide the denoising process. Finally, the multi-layer processed tensor is passed through an MLP decoder to obtain the denoised action sequence $\hat{A}^{(0)}$, which are further transformed into generated denoised trajectories via a dynamics model $f$.

\textbf{Trajectory predictor and score predictor}: These two prediction modules serve as auxiliary training tasks to enhance the model's understanding of the scenario and the quality of generated trajectories. Both modules share partial network structures and take the scenario context $\hat{\boldsymbol{c}}$ and a predefined set of $Q$ candidate target points as input. Specifically, the candidate target points are first encoded through an MLP, then fed into a Transformer network containing multiple cross-attention blocks to interact with the scenario context $\hat{\boldsymbol{c}}$, and finally processed by different decoders to generate their respective outputs.

For the score predictor, the goal is to assess the plausibility of each candidate target point as a potential future target for a specific agent. It generates two critical scores: 1) the probability score $P_{score} = [p_1^{1:Q},..., p_i^{1:Q}, ..., p_N^{1:Q}]$, which indicates the likelihood of the $Q$ candidate points being selected as the optimal target point $q^*$ of each agent $i$, 2) the drivability score $D_{score} = [d_1^{1:Q},..., d_i^{1:Q}, ..., d_N^{1:Q}]$, which evaluates the $Q$ candidate points' feasibility for safe and effective navigation of each agent $i$. The trajectory predictor forecasts the action sequences $A_{pred} = [A_1^{1:Q}, A_2^{1:Q}, ..., A_N^{1:Q}]$ required for all agents to reach each candidate target point, where $A_i^{1:Q}$ represents the set of $Q$ action sequences predicted for the $i$-th agent. After that, the predicted action sequences $A_{pred}$ are subsequently transformed through the dynamics model $f$ to generate all predicted trajectories $T_{pred} = [T_1^{1:Q}, T_2^{1:Q}, ..., T_N^{1:Q}]$.

\subsubsection{Model training}

As discussed above, to effectively train the model, we adopt a multi-task learning framework that simultaneously optimizes the main diffusion denoising task along with auxiliary trajectory prediction and score prediction tasks. The total loss function $\mathcal{L}_{\text{total}}$ is defined as the weighted sum of individual loss components, as shown in Eq. \ref{eq8}.

\begin{equation}
\label{eq8}
\begin{aligned}
    \mathcal{L}_{\text{total}} = \lambda_{\text{diff}}\mathcal{L}_{\text{diff}} + \lambda_{\text{traj}}\mathcal{L}_{\text{traj}} + \lambda_{\text{prob}}\mathcal{L}_{\text{prob}} + \lambda_{\text{driv}}\mathcal{L}_{\text{driv}}
\end{aligned}
\end{equation}

The core diffusion loss $\mathcal{L}_{\text{diff}}$ aims to minimize the L1 distance between the generated trajectories $\hat{S} $ and the ground-truth trajectories $S$:

\begin{equation}
\label{eq9}
\begin{aligned}
    \mathcal{L}_{\text{diff}} = \mathbb{E}_{S, k, \epsilon \sim \mathcal{N}(0, \textbf{I})} 
    \left[ \left\| f(\mathcal{D}_{\theta}(\tilde{A}^{(k)}, k, \hat{\boldsymbol{c}})) - S \right\|_1 \right]
\end{aligned}
\end{equation}
where $\mathcal{D}_\theta$ denotes the action denoiser, which predicts the noise-free action sequence $\hat{A}^{(0)}$ based on the noised action sequence $\tilde{A}^{(k)}$, noise level $k$, and scenario context $\hat{\boldsymbol{c}}$. The predicted actions are then further transformed through the dynamics model $f$ to derive the predicted trajectory $\hat{S} = f(\hat{A}^{(0)})$.

For the auxiliary trajectory prediction loss $\mathcal{L}_{\text{traj}}$, we first compute the smooth L1 loss between the $Q$ predicted trajectories $T_i^{1:Q}$ and the ground-truth trajectory $S_i$ for each agent $i$, then select the trajectory with the minimum loss as shown in Eq. \ref{eq10}. Subsequently, the smooth L1 loss between the optimal predicted trajectories and ground-truth trajectories for all agents is averaged to obtain $\mathcal{L}_{\text{traj}}$, as formalized in \ref{eq11}.


\begin{align}
    \label{eq10}
    & T_i^* = \arg\min_{q \in \{1, \ldots, Q\}} \mathcal{L}_{\text{smoothL1}}(T_i^q, S_i) \\
    \label{eq11}
    & \mathcal{L}_{\text{traj}} = \mathbb{E}_{i\in \{1,...,N\}}\left[ \mathcal{L}_{\text{smoothL1}}(T_i^*,S_i)\right] 
\end{align}

For the auxiliary score prediction loss, it is decomposed into two components: a probability score loss $\mathcal{L}_{\text{prob}}$ and a drivability score loss $\mathcal{L}_{\text{driv}}$. The probability score loss aims to guide the model to assign the highest probability to the candidate goal point $q_i^*$ corresponding to the optimal trajectory. It computes the cross-entropy loss between the $Q$ probability scores $p_i^{1:Q}$ output by the score predictor and a one-hot vector $y_{wta}^i$ representing the index of $q_i^*$, and then averages the loss over all agents, as shown below.

\begin{equation}
\label{eq12}
\begin{aligned}
    \mathcal{L}_{\text{prob}} = \mathbb{E}_{i\in\{1,...,N\}}\left[\text{CrossEntropy}(y_{wta}^i, p_i^{1:Q})\right]
\end{aligned}
\end{equation}

The drivability score loss $\mathcal{L}_{\text{driv}}$ evaluates whether the model correctly predicts the drivability of each candidate goal point. Let $y_{driv}^{q,i}\in\{0,1\}$ denote the ground-truth label indicating whether the $q$-th candidate point of agent $i$ lies within a drivable area, which determined by extracted map data. $\mathcal{L}_{\text{driv}}$ first computes the binary cross-entropy between $y_{driv}^{q,i}$ and $d_i^q$ (the predicted drivability of agent $i$'s $q$-th candidate target point), then averages the loss over all $Q$ candidate points, and finally obtains the average drivability prediction loss for all agents as shown in the following equation:

\begin{equation}
\label{eq13}
\begin{aligned}
    \mathcal{L}_{\text{driv}} = \mathbb{E}_{i\in\{1,...,N\}}\mathbb{E}_{q\in\{1,...,Q\}}\left[\text{BCEWithLogits}(y_{driv}^{q,i},d_i^q)\right]
\end{aligned}
\end{equation}

The key hyperparameters and configurations used during model training are summarized in Table \ref{tab1}.

\begin{table}[width=\linewidth,cols=3,pos=t]
\renewcommand{\arraystretch}{1.3}
\caption{Hyperparameters and configurations used during model training.}
\label{tab1}
\begin{center}
\begin{tabular}{ m{0.2\linewidth} m{0.18\linewidth}  m{0.15\linewidth}}
\toprule
\textbf{Parameter type} & \textbf{Parameter Name} & \textbf{Value}  \\
\midrule
\multirow{3}{\linewidth}{Optimizer related} & Optimizer type & AdamW \\
~ & Learning rate & 2e-4 \\ 
~ & Weight decay & 0.01 \\ \hline
\multirow{2}{\linewidth}{Training process related} & Batch size & 6 \\
~ & Epochs & 15 \\ \hline
\multirow{2}{\linewidth}{Diffusion model related} & Total diffusion step $K$ & 50 \\
~ & Noise schedule type & Cosine \\ \hline
\multirow{2}{\linewidth}{Weights of loss functions} & $\lambda_{\text{diff}}$ and $\lambda_{\text{traj}}$ & 1.0 \\
~ & $\lambda_{\text{prob}}$ and $\lambda_{\text{driv}}$ & 0.5 \\ \hline
Hardware related & GPU & NVIDIA 4090 $\times \ 8$ \\

\bottomrule
\end{tabular}
\end{center}
\end{table}

\subsection{Adaptive guided diffusion}

\subsubsection{Adaptive guidance framework and algorithm}

Generally, the guidance process is implemented by intervening in the sampling steps of the reverse denoising process. More in detail, at each reverse denoising step $k$, the predicted mean $\mu_k$ in Eq. \ref{eq5} is modified based on a differentiable guidance function $\mathcal{G}(\cdot)$, thereby enabling the scenario generation task toward desired directions, as formalized below \citep{zhong2022guided}:
\begin{equation}
\label{eq14}
\begin{aligned}
    p_\theta(\tilde{A}^{(k-1)} \vert \tilde{A}^{(k)}, \boldsymbol{c}) : = \mathcal{N}(\tilde{A}^{(k-1)}; \mu_k - \lambda \sigma^2_k\nabla_{\mu_k} \mathcal{G}, \sigma^2_k\textbf{I}) 
\end{aligned}
\end{equation}
where $\lambda$ is a hyperparameter that controls the strength of guidance.

In particular, instead of employing static fixed guidance strategies, we introduce time domain and trigger distance to dynamically activate different guidance functions, thereby achieving adaptive guidance based on the VUT's responses and scenario dynamics. As shown in Fig. \ref{Framework}, the time domain $[t_{s,j}, t_{e,j}]$  defines the valid window of the guidance function $\mathcal{G}_j$ on the simulation timeline, which controls the effective time period for guidance activation. The trigger distance defines the scenario requirements that must be satisfied for the guidance function to take effect. Even within the valid time domain, the corresponding guidance gradient is computed and applied only when the trigger distance is met.

Additionally, during the computation of $\mathcal{G}$ which requires trajectory-level information as input, we directly utilize $\mu_k$ as the noised action sequence to derive trajectories (namely $\mathcal{G}[f(\mu_k)]$), thereby avoiding repeated calls to the computationally intensive action denoiser $\mathcal{D}_\theta$ and improving computational efficiency. However, this approach exhibits limitations in guidance stability compared to invoking $\mathcal{D}_\theta$ at each guidance step $i$ to obtain noise-free actions before trajectory computation (namely $\mathcal{G}\{f[\mathcal{D}_\theta(\mu_k,k,\hat{\boldsymbol{c}})]\}$). As a remedy, our method initiates guidance only at step 
$K_{\text{guide\_start}}$ $(K\ge K_{\text{guide\_start}}\ge 1)$  of the denoising process, by which point $\mu_k$ has already approached noise-free actions due to prior denoising stages. The adaptive guidance diffusion algorithm is presented in Algorithm \ref{alg1}.

\begin{algorithm}[t]
\caption{The adaptive guidance diffusion algorithm}
\label{alg1}
\SetAlgoLined
\KwIn{
    Total simulation timesteps $T_{sim}$, total diffusion steps $K$, number of guidance iterations $N_{\text{guide}}$, replan frequency $f_{\text{rep}}$, time domain $[t_{s,j}, t_{e,j}]$, set of guidance functions $\{\mathcal{G}_j\}$ and their weights $\{\omega_j\}$, scenario context $\boldsymbol{c}$, well-trained scenario encoder $\mathcal{E}_{\theta}$ and action denoiser $\mathcal{D}_\theta$, noise schedule parameters $\alpha_k$, $\bar{\alpha}_k$, $\beta_k$, dynamics model $f$
    }

\For{$t_{sim}=0\ \text{to}\ T_{sim}$}{
    $\text{flag} \gets t_{sim} \mod f_{\text{rep}}$ \;
    \tcc{Determine whether replanning is required}
    \If{\text{flag} == 0}{
        $\hat{\boldsymbol{c}} \gets \mathcal{E}_{\theta}(\boldsymbol{c})$ \;
        $\tilde{A}^{(K)} \sim \mathcal{N}(0, \mathbf{I})$  \Comment*{Sample initial pure noise actions}
        \For{$k=K$ to $1$}{
            $\hat{A}^{(0)} \gets \mathcal{D}_\theta(\tilde{A}^{(k)}, k, \hat{\boldsymbol{c}})$ \Comment*{Predict noise-free actions}
            $\mu_k \gets \frac{\sqrt{\bar{\alpha}_{k-1}} \beta_k}{1 - \bar{\alpha}_k} \hat{A}^{(0)} + \frac{\sqrt{\alpha_k} (1 - \bar{\alpha}_{k-1})}{1 - \bar{\alpha}_k} \tilde{A}^{(K)}$ \Comment*{Calculate the sampling mean (Eq.\ref{eq6})}
            \tcc{Determine whether guidance is required}
            \eIf{$k \le K_{\text{guide\_start}}$}{
                $\mu_k' \gets \mu_k$  \;
                \For{$i=1$ to $N_{\text{guide}}$}{
                    $S_k' \gets f(\mu_k')$ \Comment*{Directly compute the noised trajectories}
                    $\mathcal{G}_{\text{total}} \gets \sum_j \omega_j \mathcal{G}_j(S_k', \mu_k', \hat{\boldsymbol{c}} ) \mathbb{1} (t_{sim}\in[t_{s,j}, t_{e,j}])$ \Comment*{Calculate the sum of guidance function values within the valid time domain}
                    $\mu_k'\gets \mu_k' - \lambda \sigma^2_k\nabla_{\mu_k'} \mathcal{G}_{\text{total}}$ \Comment*{Optimization step}
                }
            }
            {
                $\mu_k' \gets \mu_k$\;
            }
            $\sigma^2_k = \frac{1-\bar{\alpha}_{k-1}}{1-\bar{\alpha}_k} \beta_k$ \Comment*{Calculate the sampling variance (Eq.\ref{eq7}) }
            $\epsilon_k \sim \mathcal{N}(0, \mathbf{I})$\;
            $\tilde{A}^{(k-1)} \gets \mu_k' + \sqrt{\sigma_k^2}\epsilon_k$ \;
            
        }
        
    }
    $\text{Simulate}(\tilde{A}^{(0)})$\;
    $\text{Update}(\boldsymbol{c})$\;
}
\end{algorithm}

\subsubsection{Guidance functions} \label{sec3-3-2}

To achieve flexible adversarial guidance, a series of basic guidance function templates are predefined. These templates enable the VLM to selectively combine them for complex guidance objectives, while avoiding code bugs that may arise from direct code generation by the VLM.

\textbf{The control smoothness guidance function} $\mathcal{G}_{\text{smooth}}$ is a global guidance function that operates throughout the entire simulation process to penalize abrupt control actions, thereby addressing potential stability and consistency challenges in guidance. Specifically, its objective is to minimize the weighted sum of squares of acceleration and yaw rate in the target agent's action sequence, as shown in the following equation:

\begin{equation}
\label{eq15}
\begin{aligned}
    \mathcal{G}_{\text{smooth}} = \sum_{t=t_{sim}+1}^{t_{sim}+T_{fut}}\sum_{i=1}^N \left( \omega_{\text{acc}} \cdot (\hat{a}_{\text{acc},t}^i)^2 + \omega_{\text{yaw}} \cdot (\hat{a}_{\text{yaw}, t}^i)^2\right)
\end{aligned}
\end{equation}
where $a_{\text{acc},t}^i$ and $a_{\text{yaw}, t}^i$ represent the acceleration and yaw rate in the predicted action of agent $i$ at timestep $t$, and $\omega_1$ and $\omega_2$ are the corresponding weights. $\mathcal{G}_{\text{smooth}}$ is typically combined with other guidance functions mentioned below, serving as a regularization term to ensure that while pursuing specific objectives, the generated trajectories maintain fundamental smoothness and comfort.

\textbf{The speed guidance function} $\mathcal{G}_{\text{speed}}$ is used to guide specific agent to travel at a speed close to a target velocity $v_{\text{target}}$. It calculates the mean absolute deviation between the agent's predicted velocity $\hat{v}_t$ and the target velocity within the predicted trajectory. To ensure that smaller deviations yield higher rewards, the negative value is used as shown in the following equation:
\begin{equation}
\label{eq16}
\begin{aligned}
    \mathcal{G}_{\text{speed}} = -\frac{1}{T_{fut}}\sum_{t=t_{sim}+1}^{t_{sim}+T_{fut}} |\left\| \hat{v}_t \right\|_2-v_{\text{target}} |
\end{aligned}
\end{equation}

\textbf{The goal point guidance function} $\mathcal{G}_{\text{goal}}$ is designed to direct the specified agent to a predefined target point $p_{\text{goal}}$. It calculates the minimum distance between the agent's predicted trajectory positions $p_t$ and the target point $p_{\text{goal}}$ over the prediction horizon $[t_{sim}+1:t_{sim}+T_{fut}]$. Similar to Eq. \ref{eq16}, the negative value of this minimum distance is used in Eq. \ref{eq17} to ensure that smaller distances yield higher rewards:

\begin{equation}
\label{eq17}
\begin{aligned}
    \mathcal{G}_{\text{goal}} = - \min_{t\in [t_{sim}+1,t_{sim}+T_{fut}]} \mathcal{L}_{\text{dist}}(p_t,p_{\text{goal}})
\end{aligned}
\end{equation}
where $\mathcal{L}_{\text{dist}}$ denotes the distance metric function.

\textbf{The interaction guidance function} $\mathcal{G}_{\text{interact}}$ is designed to encourage specified agents to continuously challenge the VUT in testing scenarios, thereby revealing the VUT's flaws and limitations. It calculates the minimum distance between all the specified agents and the VUT over the prediction horizon $[t_{sim}+1:t_{sim}+T_{fut}]$. By minimizing this value, it increases the probability of close-range interactions between agents and the VUT, as formalized below:

\begin{equation}
\label{eq18}
\begin{aligned}
    \mathcal{G}_{\text{interact}} = - \min_{t\in [t_{sim}+1,t_{sim}+T_{fut}]} \min_{i \in \mathcal{A}_{\text{adv}}} \|p_t^i, p_t^{\text{VUT}} \|_2  \mathbb{1}(\|p_{t_{sim}}^i, p_{t_{sim}}^{\text{VUT}} \|_2 \le d_{\text{trigger}})
\end{aligned}
\end{equation}
where $\mathcal{A}_{\text{adv}}$ denotes the set of specified adversarial agents, $p_{t_{sim}}^i$ and $p_{t_{sim}}^{\text{VUT}}$ represent the predicted positions of agent $i$ and the VUT at timestep $t$, respectively. $\mathbb{1}(\cdot)$ is an indicator function implementing the trigger condition mechanism described earlier: it ensures that the interaction guidance function is computed only when the current actual distance between the agent and the VUT is below a predefined threshold $d_{\text{trigger}}$. When the actual distance exceeds $d_{\text{trigger}}$, the guidance value remains 0 even within the valid time domain. This mechanism allows agents to adaptively determine when to initiate adversarial interactions based on real-time scenario states, while avoiding unnatural strong guidance behavior when the vehicles are still far apart.

Finally, the total guidance function $\mathcal{G}_{\text{total}}$ can be computed as a weighted sum of individual guidance functions, as formalized below:
\begin{equation}
\label{eq19}
\begin{aligned}
    \mathcal{G}_{\text{total}} = \sum_{j=1}^J \omega_j \cdot \mathcal{G}_j \mathbb{1}(t_{sim}\in [t_{s,j}, t_{e,j}])
\end{aligned}
\end{equation}
where $\omega_j$ is the weight of the $j$-th guidance function, which is determined by VLM in this paper. The indicator function $\mathbb{1}(\cdot)$ implements the time domain mechanism described earlier, ensuring that the corresponding guidance function value is computed only within the valid time domain.

\subsection{VLM-driven guidance generation} \label{sec_vlm}
As discussed in Section \ref{sec1}, this paper treats the VLM as a high-level strategist that determines scenario generation objectives and formulates guidance functions, thereby achieving VLM-directed safety-critical testing scenario generation. Compared to traditional user-specified approaches where LMs primarily serve as translators, our method fully leverages the VLM's significant potential in scenario understanding and reasoning. The proposed method mitigates limitations arising from the finiteness and subjectivity of expert knowledge while simultaneously improving scenario generation efficiency.

\subsubsection{VLM inputs}
\begin{figure}[pos=b] 
      \centering
      \includegraphics[width=16cm]{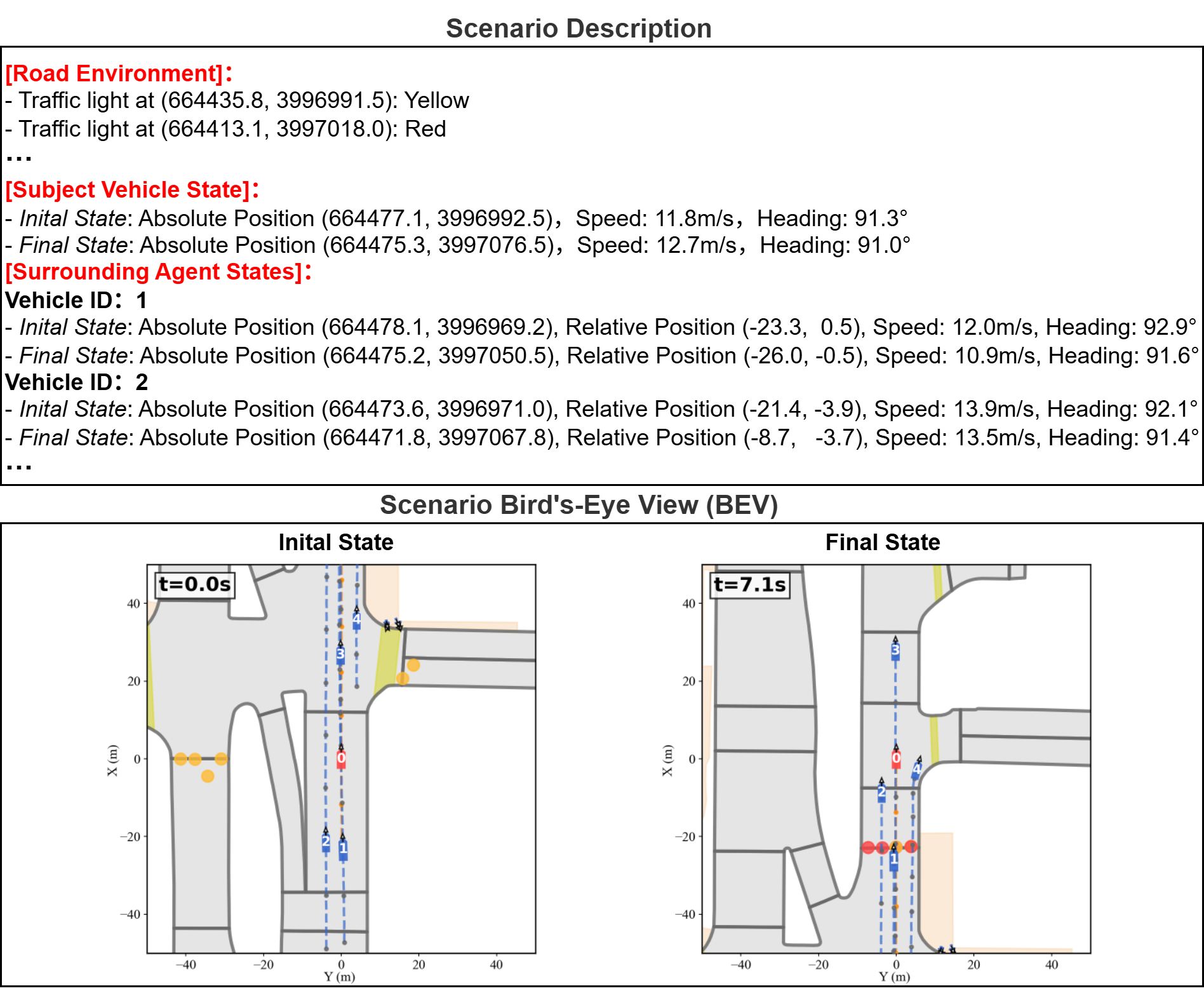}
      \caption{An example of the VLM input information.}
      \label{VLM_input}
\end{figure}

Compared to LLMs that rely solely on textual input, VLMs can leverage multimodal inputs to achieve more comprehensive scenario understanding. Specifically, the multimodal input for the VLM in this paper includes two components: scenario textual descriptions and a BEV image of the scenario, as demonstrated in Fig. \ref{VLM_input}.

Scenario description information is extracted from simulation logs in the original dataset and converted into structured information that is easy to understand and process. It includes both static environmental data and dynamic agent states. Due to the temporal dependencies between sequential agent states, to avoid input redundancy, the scenario description only contains information related to the initial and final states of the scenario. Particularly, for surrounding agent states, the scenario description includes not only their absolute positions but also their relative positions to the SV. This SV-centric perspective facilitates the subsequent generation of critical scenarios targeting the SV.

Similar to the scenario description, the scenario BEV includes only two images representing the initial and final states of the scenario. These two images are displayed side-by-side to highlight the differences between the two states, aiding the VLM in understanding the temporal evolution of the scenario. More in detail, each image is consistently centered on the SV and rotated such that the SV’s initial heading is oriented toward the top of the image. Meanwhile, the coordinate axes are defined with the SV as the origin. This layout facilitates the VLM in quickly locating the SV and accurately estimating the position and distance of surrounding agents relative to the SV. Additionally, as an image, the scenario BEV contains richer information compared to scenario textual descriptions, including (1) Vehicle geometry and trajectory information: the BEV depicts all vehicles' current positions, shapes, IDs, orientations, as well as their trajectories throughout the scenario; (2) Map and environmental elements: the BEV also contains lane markings, crosswalks, and traffic lights, providing the VLM with crucial environmental constraints and traffic rule information.

In summary, the scenario BEV provides intuitive perception of the spatial layout and dynamic processes, while the scenario textual description offers precise quantitative data and semantic information. The two modalities complement each other and jointly support the VLM in subsequent tasks such as scenario understanding, criticality analysis, and reasoning.

\subsubsection{Accident knowledge database}

To enable the VLM to better understand the essence of safety-critical scenarios, thereby correctly performing criticality analysis and potential risk identification, and subsequently generating realistic and reasonable safety-critical testing scenarios, this paper utilizes real-world accident data as prior knowledge to guide the VLM. Specifically, the accident knowledge database constructed in this paper is derived from the crash scenario typology report published by the National Highway Traffic Safety Administration (NHTSA) \citep{najm2007precrash}. This report defines 37 typical collision scenarios based on the analysis of light vehicle accident data.

\begin{table}[width=\linewidth,cols=4,pos=b]
\renewcommand{\arraystretch}{1.3}
\caption{Accident knowledge database that associates SV behaviors with typical accident scenarios.}
\label{tab2}
\begin{center}
\begin{tabular}{ m{0.12\linewidth} m{0.17\linewidth}  m{0.18\linewidth}  m{0.4\linewidth}}
\toprule
\textbf{SV behaviors} & \textbf{Conflict vehicle} & \textbf{Adversarial behavior} & \textbf{Typical accident scenario descriptions}  \\
\midrule
\multirow{3}{\linewidth}{Going straight} & Adjacent vehicle & Cut-in  & A vehicle in the adjacent lane suddenly cuts in front of the SV. \\
~ & Leading vehicle & Emergency braking  & The leading vehicle of the SV brakes abruptly without warning. \\ 
~ & Cross-traffic vehicle & Red-light running  & A vehicle traveling laterally through the intersection runs a red light. \\ \hline
Turning & Oncoming vehicle & Trajectory conflict & The oncoming through vehicle fails to yield to the SV making a left turn.  \\ \hline
\multirow{2}{\linewidth}{Lane changing} & Occluded vehicle & Collision & A rapidly approaching vehicle is located in the blind spot of the SV's target lane. \\
~ & Adjacent vehicle & Trajectory interference  & A vehicle in the adjacent lane performs a simultaneous lane change maneuver. \\ \hline
\multirow{2}{\linewidth}{Ramp merging} & Main road vehicle & Rear-end collision & The SV is rear-ended by a main-road vehicle while merging due to a speed differential. \\
~ & Ramp vehicle & Forcible merging & A vehicle from the ramp executes a forced merge in front of the SV. \\ 

\bottomrule
\end{tabular}
\end{center}
\end{table}
Generally, a SV in a specific scenario tends to have a clear driving intention (e.g., turning left at an intersection, proceeding straight, etc.) and exhibits corresponding driving behaviors. This characteristic provides a basis for the VLM to perform risk association, meaning that the SV's driving behavior can serve as a key reference for classifying safety-critical scenarios and determining scenario generation objectives. Based on the above analysis, this paper constructs a accident knowledge database linking SV behaviors to common accident types, as shown in Table \ref{tab2}.

\subsubsection{CoT-based generation objective determination and guidance function formulation}

\begin{figure}[pos=b] 
      \centering
      \includegraphics[width=\linewidth]{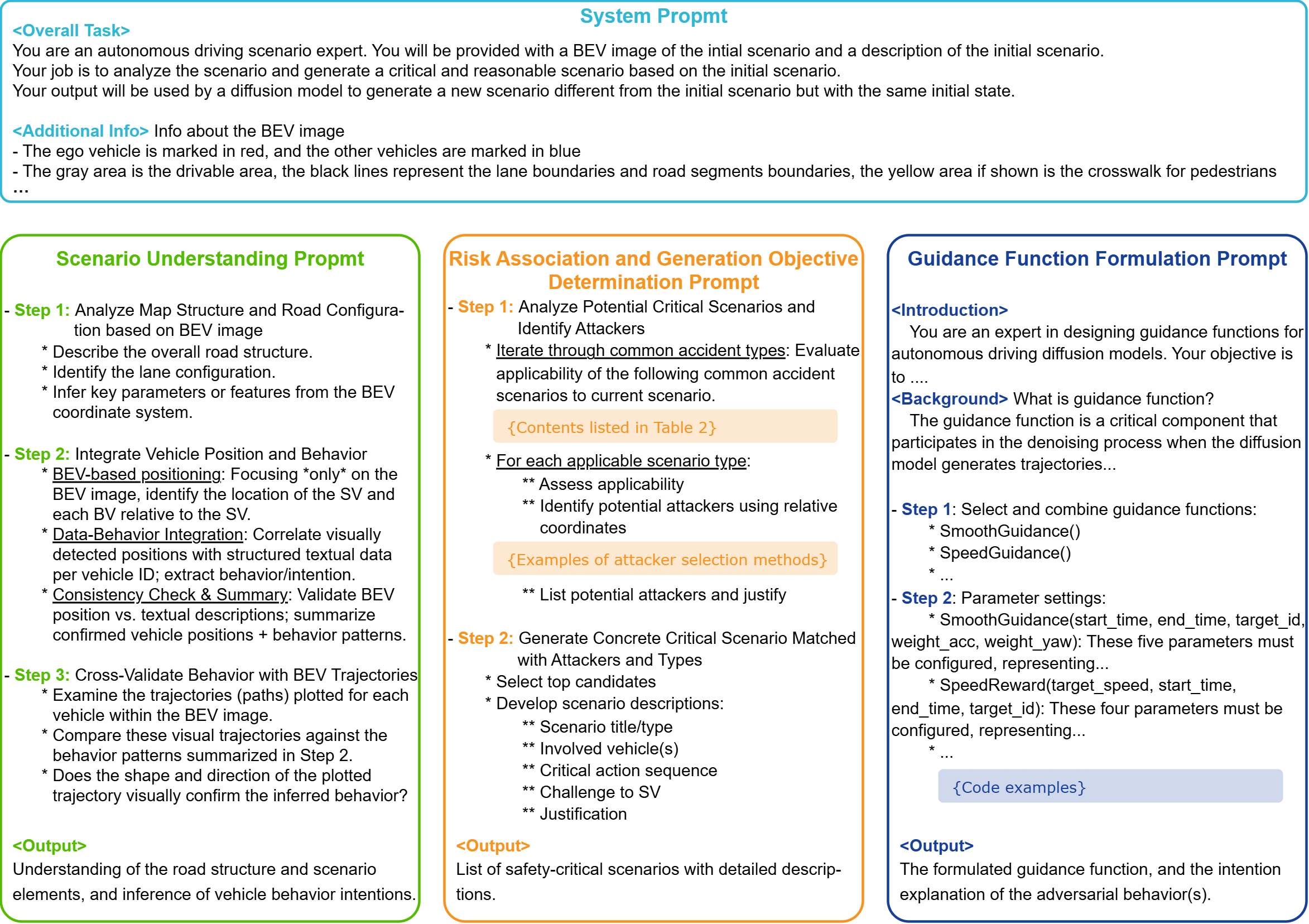}
      \caption{The prompts for each CoT reasoning step.}
      \label{VLM_prompt}
\end{figure}

In this paper, instead of having the VLM directly generate the guidance function in a single step after receiving the input, we introduce a chain-of-thought (CoT) reasoning process \citep{wei2022chainofthought} to enhance its reasoning capability on complex tasks. Concurrently, we require the VLM to output intermediate reasoning results at each step, thereby improving the interpretability of the reasoning process. The CoT reasoning process comprises three steps: (1) Scenario understanding, (2) Risk association and generation objective determination, and (3) Guidance function formulation. The prompts for each step in the CoT reasoning process are illustrated in Fig. \ref{VLM_prompt}. As shown in the figure, besides the three reasoning stages mentioned above, a system prompt is first designed to define the overall task of the VLM and provide additional information about the BEV image. For ease of understanding, we have omitted the example explanation section after each prompt and simplified the original prompts.

\textbf{Scenario understanding prompts} are used to structurally guide the VLM in conducting an in-depth analysis of the scenario. 
Based on the BEV images, the map structure and road configuration of the scenario are first identified. Subsequently, combined with the scenario textual description, the positions and ID information of all vehicles are extracted, and their driving intentions are recognized. After that, the trajectory information in the images is used to cross-validate the identified driving intentions. Finally, the VLM is required to output its understanding of the scenario and the recognition results of the driving intentions.

\textbf{Risk association and generation objective determination prompts} are designed to guide the VLM in linking its understanding of the scenario with the established accident knowledge database, actively associating the current situation with potential critical scenario types it may evolve into. Using the identified SV behavior in the previous step as an index, the VLM queries the accident scenario types in Table \ref{tab2} and comprehensively evaluates the likelihood of the current scenario evolving into the corresponding accident scenario. It then further determines the adversarial vehicle and its behavioral intention to generate a concrete safety-critical scenario description.

\textbf{The guidance function formulation prompts} are used to guide the VLM in translating the aforementioned high-level analysis results into executable guidance functions code at the underlying level and configuring their parameters. To enhance the usability of the VLM's output code, we predefine a series of guidance function templates based on Section \ref{sec3-3-2} and provide the VLM with detailed explanations of the specific meanings of each function parameter. Based on the scenario analysis results, the VLM selects and combines guidance functions while configuring parameters such as their time domain, trigger distance, target vehicles, and weights. The output guidance function code will be directly used for guided diffusion generation in the operational layer.

\section{Experiments} \label{sec4}

In this section, we apply the proposed method in practical applications. We first introduce the dataset, simulation platform, the algorithms used to control the VUT, and the evaluation metrics. Subsequently, we demonstrate the overall performance of the proposed method in generating safety-critical testing scenarios. After that, through the analysis of multiple practical cases, we validate the proposed method's adaptability across different scenarios and various AUTs, and the performance in VLM-directed generation. Finally, an ablation study is conducted to investigate the impact of different input modalities and prompting strategies on the VLM's comprehension capabilities. 

It is noteworthy that the core innovation of this paper lies in the VLM-directed scenario generation concept and its corresponding three-layer hierarchical framework for adaptive scenario guidance. Any other state-of-the-art (SOTA) foundational diffusion models and guided diffusion methods can be integrated into this framework. Therefore, this paper does not select baselines for comparison.

\subsection{Dataset and simulation platform}

The experiments in this paper are conducted on the nuPlan dataset \citep{caesar2022nuplan}, which primarily contains approximately 1500 hours of driving logs from four cities: Las Vegas, Pittsburgh, Boston, and Singapore. These driving logs encompass diverse road structures, traffic rules, and driving styles, forming a rich and varied library of driving scenarios. 
To efficiently utilize the nuPlan dataset for model training and testing, we first preprocess the raw data into scenario segments with a duration of approximately 10 seconds, which are further divided into historical and future parts. In this paper, $T_{hist}=11$ and $T_{fut} = 81$, with an interval of 0.1 seconds between time steps. As mentioned earlier, during model training, the historical trajectory serves as the model input to encode the past state of the scenario, while the future trajectory acts as the supervision signal to compute the model's prediction loss and guide the model toward accurate predictions.

\begin{figure}[pos=t] 
      \centering
      \includegraphics[width=.75\linewidth]{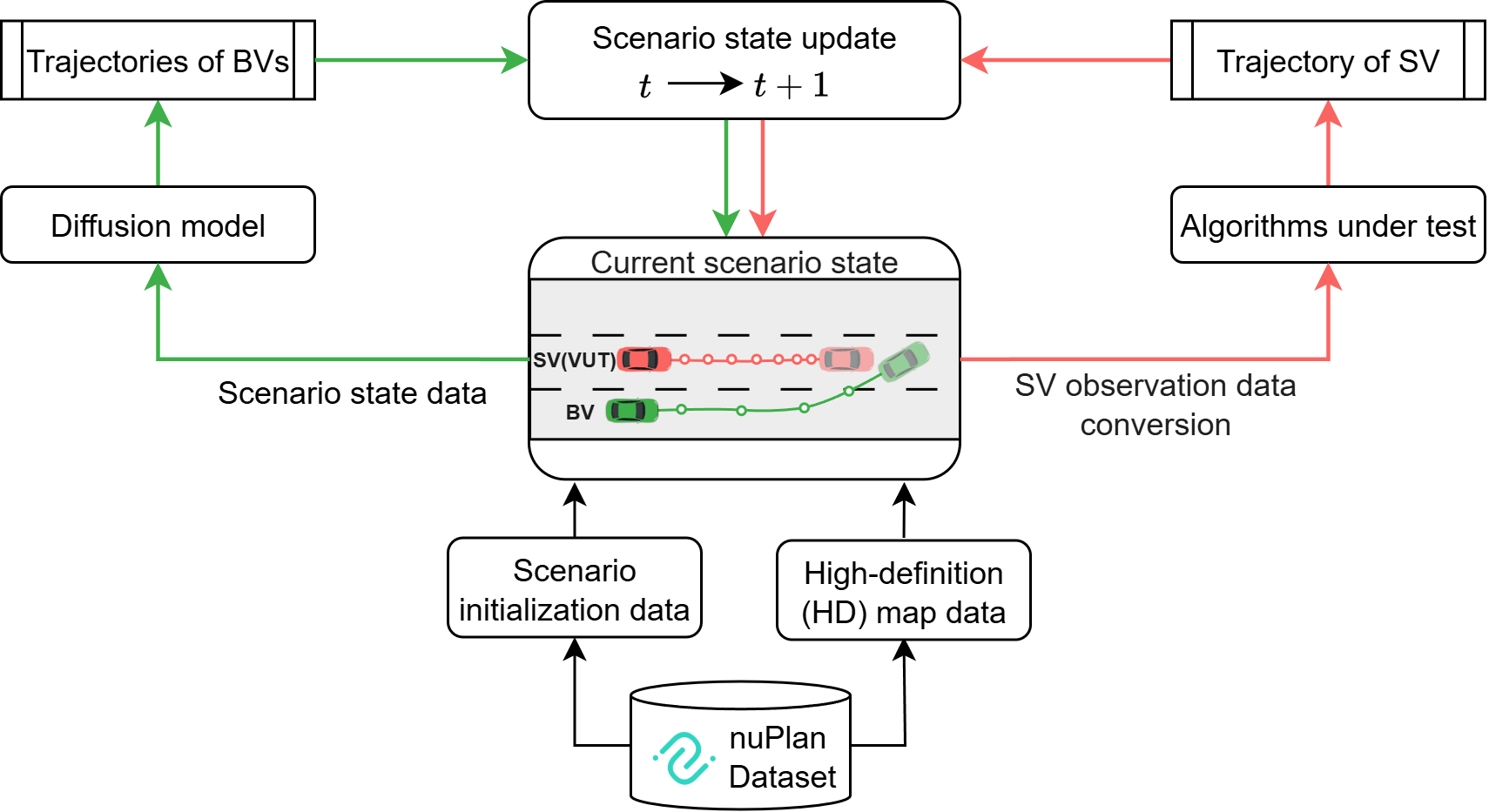}
      \caption{The co-simulation platform.}
      \label{sim_platform}
\end{figure}

As for simulation platform, most existing research based on the nuPlan dataset utilizes the official nuPlan simulation development kit (nuplan-devkit) as the foundational platform. However, since its native simulation architecture is primarily designed around the planning and control of the SV, its simulation of BVs' behavior is relatively simplified. This makes it difficult for the simulation platform to integrate external models for controlling BVs or to achieve selective control of specific BV. As a remedy, this paper designs a lightweight co-simulation platform as illustrated in Fig. \ref{sim_platform}.

As illustrated in Fig. \ref{sim_platform}, the core design concept of the simulation platform is to drive vehicle state updates through trajectories, enabling simultaneous closed-loop control of BVs via diffusion model and SV through AUTs. Meanwhile, the simulation platform achieves selective vehicle control through vehicle IDs, enabling three types of vehicles to coexist simultaneously in the simulated environment: (1) the SV controlled by AUTs, (2) specific BV(s) governed by the diffusion model, and (3) other BVs following their original trajectories. Furthermore, the platform achieves frequency decoupling between these two closed-loop control processes. In this paper, the SV performs state updates at 10Hz, while diffusion model generates and updates BV trajectories at 1Hz. This design significantly enhances the simulation platform's adaptability to AUTs with varying output frequencies.

\subsection{Experimental setup}

\subsubsection{Algorithms under test (AUTs)}
To comprehensively evaluate the performance of the generated safety-critical testing scenarios, this paper selects multiple  representative autonomous driving D\&P algorithms for testing. These algorithms cover diverse design concepts and technical approaches, aiming to verify the adaptability of our proposed method across different algorithm types. The selected algorithms are as follows:

(1) \textit{UrbanDriver}: UrbanDriver \citep{scheel2022urban} is a learning-based decision-making and planning algorithm that employs an off-policy imitation learning approach with policy gradient optimization. It trains the policy network to minimize the discrepancy between predicted trajectories and expert demonstrations. By leveraging differentiable simulation for efficient training, the method implicitly captures and mimics the diverse behaviors and interaction patterns exhibited by human drivers in complex urban scenarios. This enables the generation of naturalistic and adaptive driving policies that capable of handling intricate real-world environments. 

(2) \textit{Predictive Driver Model-Closed (PDM-Closed)}: PDM-Closed \citep{dauner2023parting} is a typical optimization-based decision-making and planning algorithm. It extends the classical Intelligent Driver Model (IDM) by incorporating Model Predictive Control (MPC) mechanisms such as environmental forecasting and trajectory scoring. This enhancement maintains real-time performance while improving both safety and traffic rule compliance.

(3) \textit{Predictive Driver Model-Hybrid (PDM-Hybrid)}: PDM-Hybrid \citep{dauner2023parting} is a hybrid decision-making and planning algorithm that combines the aforementioned PDM-Closed with a purely learning-based module PDM-Open, which employs a lightweight multilayer perceptron (MLP) to directly predict future trajectories using only lane centerlines and historical vehicle states as inputs. PDM-Hybrid combines the strengths of both approaches by utilizing PDM-Closed to generate short-term safe trajectories while employing PDM-Open for long-term trajectory offset correction. This dual mechanism preserves closed-loop control stability while enhancing open-loop prediction accuracy.

(4) \textit{Goal-Conditioned Prediction to Planning (GC-PGP)}: GC-PGP \citep{hallgarten2023prediction} is a goal-oriented hybrid decision-making and planning algorithm whose core methodology consists of two stages: path-constrained candidate generation and safety-oriented trajectory decision-making. This approach can inherently inherit the prediction model's capability to model complex interactive scenarios without requiring retraining. As a hybrid algorithm combining optimization-based and learning-based methods, its advantage lies in the rule-based constraints ensuring path safety while the data-driven approach enhances scenario adaptability.

\subsubsection{Evaluation metrics}
To quantitatively evaluate the criticality of the generated scenarios and the performance of the AUT, this paper adopts a series of evaluation metrics. These metrics are designed to assess test results from multiple dimensions, as detailed below:

(1) \textit{Collision Rate} ($CR$): $CR$ measures the proportion of scenarios where collisions occur during simulation, serving as a fundamental metric for safety evaluation.

(2) \textit{At-fault Collision Rate} ($CR_{fault}$): $CR_{fault}$ quantifies the proportion of collisions directly caused by erroneous decisions or behaviors of the AUT within the total test scenarios. It exclusively counts collisions where the SV bears responsibility, thereby distinguishing whether collisions stem from algorithmic flaws or inherent scenario risks. The responsibility criteria for different collision types are specified in Table \ref{tab3}.

(3) \textit{Average Minimum Time to Collision} ($\overline{TTC}_{min}$): $\overline{TTC}_{min}$ is calculated as the average of the minimum time to collision ($TTC_{min}$) of the SV across all scenarios. A lower $\overline{TTC}_{min}$ indicates that the SV generally faces higher risks in the generated scenarios.

(4) \textit{High-Risk Exposure Rate} ($E_{highrisk}$): $E_{highrisk}$ calculates the proportion of testing scenarios where $TTC_{min}$ is below the preset safety threshold $T_\text{safety}$ (in this paper $T_\text{safety}=1s$). This metric quantifies the frequency at which the SV encounters high-risk interactions.

(5) \textit{Path Completion Rate} ($PC$): $PC$ calculates the average ratio of the SV's completed path relative to the original log at the end of each scenario. A lower PC indicates a more challenging scenario.

\begin{table}[width=\linewidth,cols=3,pos=b]
\renewcommand{\arraystretch}{1.3}
\caption{Determination of SV responsibility for different collision types.}
\label{tab3}
\begin{center}
\begin{tabular}{ m{0.15\linewidth} | m{0.3\linewidth}  m{0.35\linewidth}}
\toprule
\textbf{SV Responsibility Determination} & \textbf{At-fault collision} & \textbf{No at-fault collision}  \\
\midrule
\multirow{3}{\linewidth}{\textbf{Collision Type}} & SV collision with a static object & Collision occurring while SV is stationary  \\
~ & Frontal collision of SV & Rear collision of SV \\ 
~ & Lateral collision of SV & ~ \\ 

\bottomrule
\end{tabular}
\end{center}
\end{table}

\subsubsection{Implementation details}
During the experiments, the AUTs are first operated in the preprocessed scenarios based on the nuPlan dataset's val14benchmark subset. Then, scenarios where AUTs exhibit low collision rates yet contain sufficient interactions are filtered as the original scenarios. This step aims to ensure that the filtered original scenarios did not pose significant challenges to the AUTs, thereby enabling clearer comparison with subsequently generated safety-critical scenarios (which are created by modifying these original scenarios) to validate the effectiveness of the proposed method.

Subsequently, we extract scenario textual descriptions and BEV images from the filtered scenarios, which are then fed into the VLM for the CoT reasoning process. Finally, the guided diffusion model performs trajectory generation based on the specific guidance functions output by the VLM, generating trajectories for the selected BV(s). These trajectories are then executed in the simulation platform for real-time control, ultimately enabling closed-loop simulation through continuous interaction with the SV controlled by AUTs. This paper primarily employs Claude-3-7-Sonnet as the VLM, while also experimenting with other VLMs that will be detailed later.

\subsection{Overall performance}

The quantitative results of four AUTs across all metrics in both original and generated scenarios are presented in Table \ref{tab4}. From a testing perspective, we expect poorer performance from the AUTs, as this indicates higher scenario challenge levels. Therefore, higher collision rates and high-risk exposure rates, along with lower average minimum TTC and path completion rates observed in the generated scenarios demonstrate the effectiveness of our method. To explicitly illustrate this, we have added upward or downward arrows after each metric in Table \ref{tab4} to indicate the desired scenario generation direction.

\begin{table}[width=\linewidth,cols=9,pos=t]
\renewcommand{\arraystretch}{1.3}
\caption{Quantitative results of four AUTs across all metrics in both original and generated scenarios.}
\label{tab4}
\begin{center}
\begin{tabular}{ m{0.07\linewidth} m{0.12\linewidth} >{\centering}m{0.1\linewidth}  >{\centering}m{0.12\linewidth} m{0.01cm} >{\centering}m{0.12\linewidth}  >{\centering}m{0.12\linewidth} m{0.01cm} m{0.12\linewidth}<{\centering}} 

\toprule
\multirow{2}{\linewidth}{\textbf{AUT}} & \multirow{2}{\linewidth}{\textbf{Scenario type}} & \multicolumn{2}{c}{\textbf{Collision-related}} & ~ & \multicolumn{2}{c}{\textbf{Risk-related}} & ~ & \textbf{Task-related} \\ \cline{3-4} \cline{6-7} \cline{9-9}

~ & ~ & $CR\uparrow$ (\%) & $CR_{fault}\uparrow$ (\%) & ~ & $\overline{TTC}_{min}\downarrow$ (s) & $E_{highrisk}\uparrow$ (\%) & ~ & $PC\downarrow$ (\%) \\
\midrule
\multirow{2}{\linewidth}{PDM-Closed} & Original & 8.3 & 8.3 & ~ & 0.91 & 50.0 & ~ & 92.5  \\
~ & Generated & \textbf{66.7} & \textbf{41.7} & ~ & \textbf{0.64} & \textbf{66.7} & ~ & \textbf{66.4} \\ \hline

\multirow{2}{\linewidth}{PDM-Hybrid} & Original & 8.3 & 8.3 & ~ & 0.91 & 50.0 & ~ & 92.5  \\
~ & Generated & \textbf{83.3} & \textbf{50.0} & ~ & \textbf{0.60} & \textbf{66.7} & ~ & \textbf{62.4} \\ \hline

\multirow{2}{.8\linewidth}{GC-PGP} & Original & 25.0 & 8.3 & ~ & 0.85 & 50.0 & ~ & 77.4  \\
~ & Generated & \textbf{50.0} & \textbf{25.0} & ~ & \textbf{0.44} & \textbf{75.0} & ~ & \textbf{64.1} \\ \hline

\multirow{2}{\linewidth}{Urban Driver} & Original & 33.3 & 25.0 & ~ & \textbf{0.76} & \textbf{66.7} & ~ & 94.1  \\
~ & Generated & \textbf{83.3} & \textbf{66.7} & ~ & 0.83 & 50.0 & ~ & \textbf{72.4} \\ 

\bottomrule
\end{tabular}
\end{center}
\end{table}

As observed in Table \ref{tab4}, regarding collision-related metrics, all AUTs exhibit significantly higher collision rates ($CR$) and at-fault collision rates ($CR_{fault}$) in the generated scenarios compared to their performance in original scenarios. For instance, PDM-Closed and PDM-Hybrid show only 8.3\% $CR$ in original scenarios, but this surges to 66.7\% and 83.3\% respectively in generated scenarios. Even GC-PGP and UrbanDriver, which already demonstrate relatively higher $CR$ in original scenarios, show further deterioration in collision avoidance performance within generated scenarios. When it comes to $CR_{fault}$, the four AUTs exhibit an average $4.2\times$ increase in generated scenarios, with at-fault collisions accounting for $\ge50\%$ of all collision cases. This observation demonstrates that the majority of collisions stem directly from algorithmic decision-making flaws and performance limitations of the AUTs themselves, rather than from inherent unavoidable risks in the scenarios. Further, these findings provide empirical validation that our generated scenarios can effectively induce and expose latent safety-critical deficiencies in autonomous driving algorithms.

Moreover, risk-related metrics also confirm the increasing challenge of generated scenarios. With the exception of UrbanDriver, all other AUTs have shown significant increases in high-risk exposure rate ($E_{highrisk}$). For instance, the $E_{highrisk}$ of GC-PGP rises sharply from 50.0\% to 75.0\%, indicating more frequent involvement in low-TTC hazardous interactions. Concurrently, all algorithms except UrbanDriver exhibit lower average minimum TTC ($\overline{TTC}_{min}$) in generated scenarios, reflecting an overall risk escalation during interactions. Notably, UrbanDriver presents an inverse trend: its $E_{highrisk}$ slightly decreases (from 66.7\% to 50.0\%), while its $\overline{TTC}_{min}$ marginally improves (from 0.76s to 0.83s). This anomaly suggests that UrbanDriver’s failure mode under generated scenarios may prioritize direct collisions over prolonged low-TTC states, implying its limitations lie more in timely response capability rather than risk anticipation. 

Finally, the task-related metric path completion rate ($PC$) further corroborates the difficulty of the generated scenarios. All AUTs exhibit significant declines in $PC$ within generated scenarios. This performance degradation indicates that in the generated scenarios, the AUTs fail to complete intended driving tasks, frequently experiencing task interruptions due to collisions or encounters with high-risk events.

In summary, all tested D\&P algorithms, including optimization-based, learning-based, and hybrid approaches, exhibit significantly worse performance in collision avoidance, risk mitigation, and task completion when exposed to the safety-critical testing scenarios generated by this paper.

\subsection{Scenario adaptability study} \label{sec_versa}
\begin{figure}[pos=t] 
      \centering
      \includegraphics[width=\linewidth]{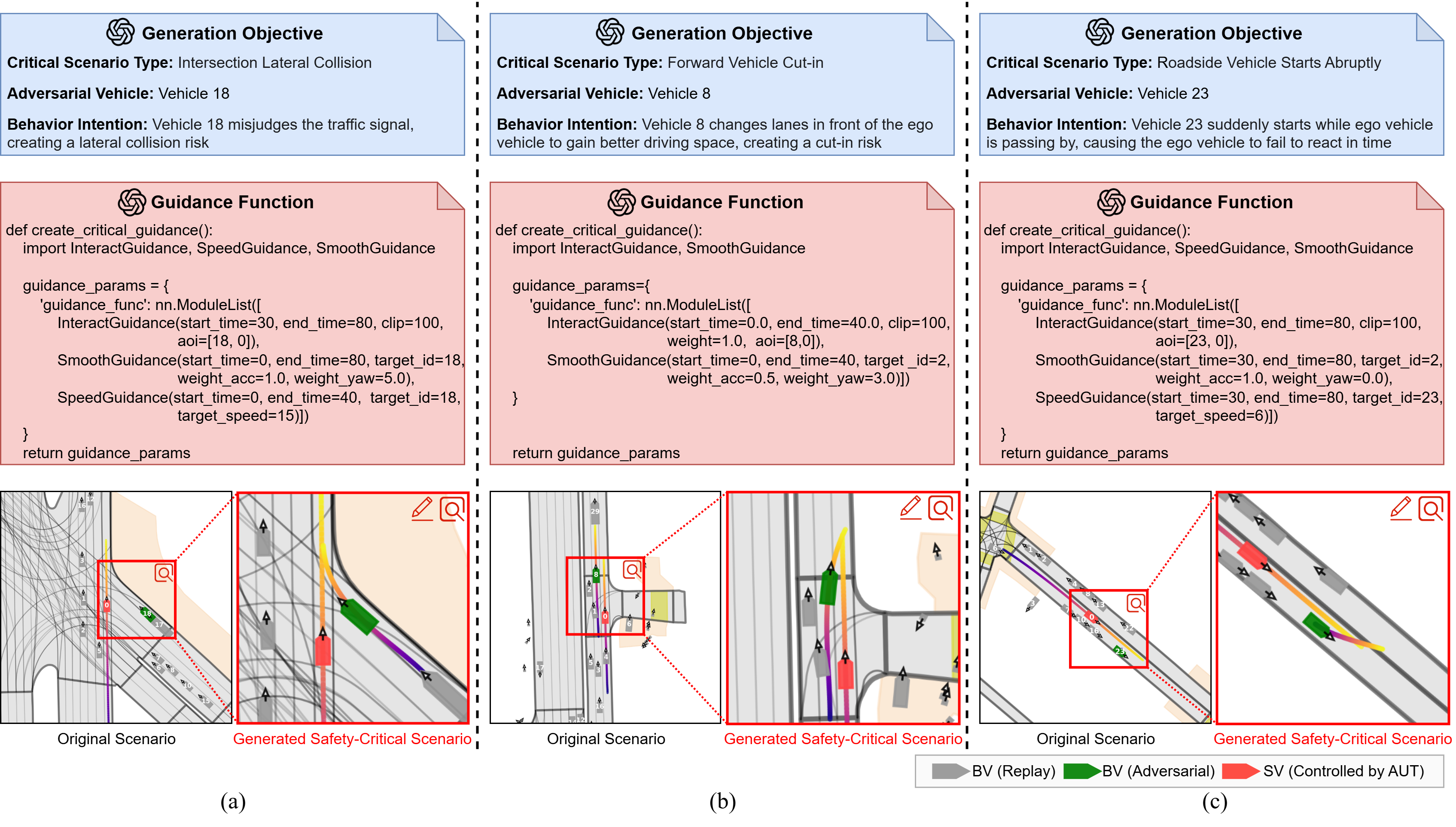}
      \caption{Examples of the generation process and results for safety-critical testing scenarios under different initial conditions.}
      \label{exp-versa}
\end{figure}

In this paper, the proposed method demonstrates strong adaptability across natural driving scenarios with diverse characteristics. Specifically, it can generate safety-critical testing scenarios that match corresponding environments and interaction patterns based on differentiated initial conditions, covering varying road structures, traffic conditions, and participant types, thereby ensuring the generalization capability of the testing approach across different scenario categories. Fig. \ref{exp-versa} illustrates examples of the generation process and results for safety-critical testing scenarios under different initial conditions. For this and subsequent case studies, PDM-Hybrid is adopted as the AUT to control the SV unless otherwise specified.

To provide a comprehensive demonstration of the entire generation process corresponding to the three-layer hierarchical framework in Fig. \ref{Framework}, Fig. \ref{exp-versa} illustrates (1) the scenario generation objectives output by the VLM (including the critical scenario type, adversarial BV ID, and adversarial behavior intention explanation), (2) the guidance functions output by the VLM (including the specific guidance function combinations and their parameters), and (3) the trajectories generated by the guided diffusion model (including both the original scenario trajectories and the generated safety-critical scenario trajectories). Additionally, in the scenario trajectory visualization, we retain only the trajectories of the SV (marked in red) and the adversarial BV (marked in green) for clarity, while omitting trajectories of other BVs (all marked in gray). For the displayed trajectories, a purple-to-gold color gradient indicates temporal progression, representing the trajectory's evolution from initial to terminal states.

Fig. \ref{exp-versa} (a) illustrates an intersection scenario where the SV initially passes through the intersection normally in the original scenario, while BV No.18 remains stationary at the traffic signal. The VLM identifies a potential lateral collision risk and initiates speed guidance to BV No.18 at the scenario's start to set it in motion from rest, followed by applying interaction guidance at the 30-th timestep to direct the BV to approach the SV as closely as possible. Consequently, the SV fails to avoid collision with the suddenly accelerating BV No.18, resulting in an accident. Fig. \ref{exp-versa} (b) and (c) demonstrate similar processes under different road topology configurations. Specifically, Fig. \ref{exp-versa} (b) shows how the VLM guides the originally straight-driving BV No.8 to perform a cut-in maneuver against the SV, while Fig. \ref{exp-versa} (c) illustrates the VLM guiding the parked BV No.23 to abruptly start moving, thereby challenging the SV. 

By observing the specific guidance function parameters, we can find that the VLM dynamically adjusts these parameters based on different scenario conditions. For instance, in Fig. \ref{exp-versa} (b), the VLM applies speed guidance to the BV from the very beginning of the scenario, enabling the BV to approach the SV earlier. While in Fig. \ref{exp-versa} (c), the VLM maintains the BV in a stationary state initially, only applying speed guidance at the 30-th timestep after the SV has approached the BV. 
The three case studies above provide clear evidence that our method can effectively identify and extract potential risk information from diverse real-world driving records, and subsequently generate safety-critical scenarios with distinctive interaction patterns and high testing value, demonstrating strong scenario adaptability.

\subsection{AUT adaptability study}

\begin{figure}[pos=b] 
      \centering
      \includegraphics[width=.8\linewidth]{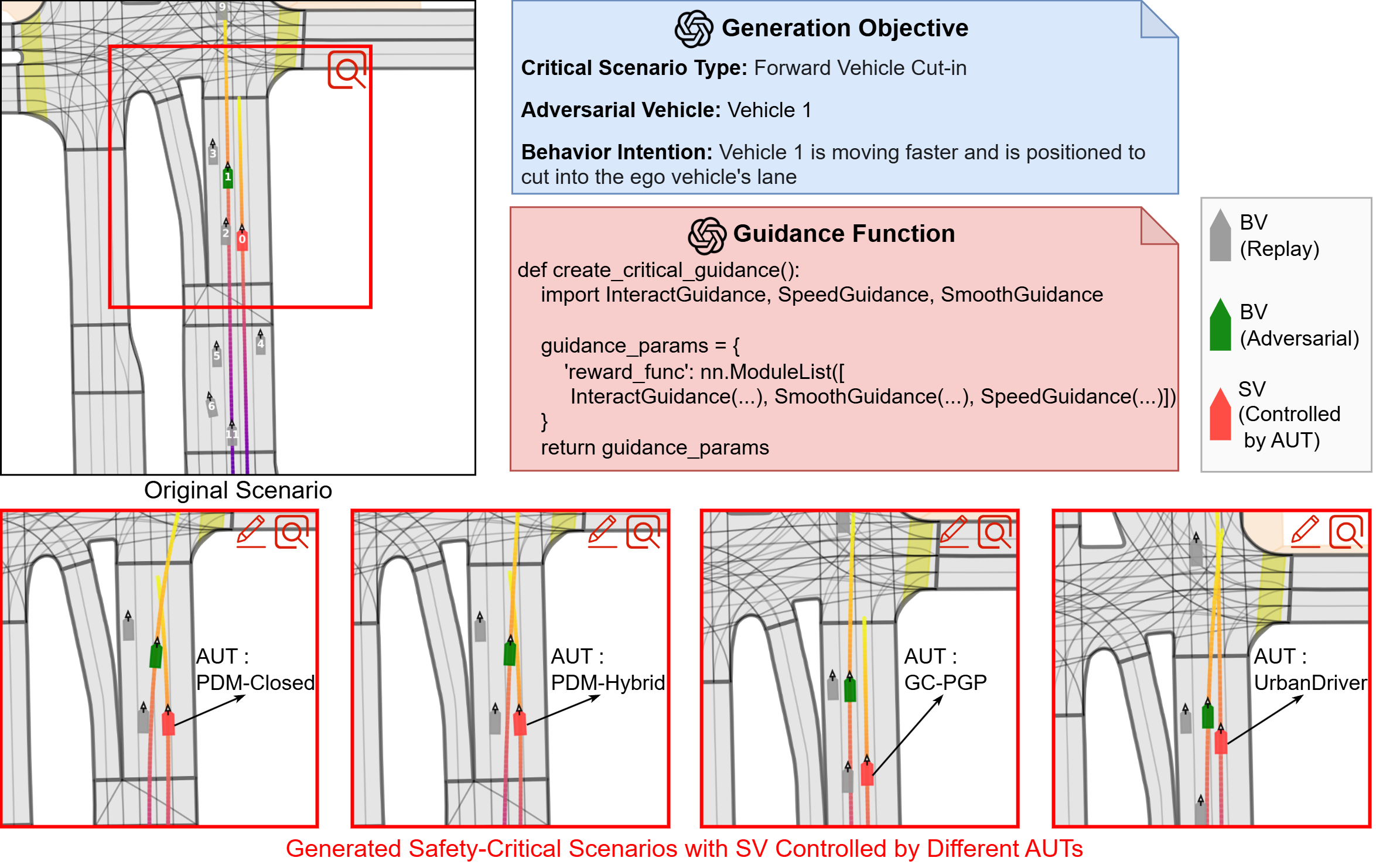}
      \caption{Examples of safety-critical scenarios generated for different AUTs under identical scenario conditions.}
      \label{exp-aut_adpt}
\end{figure}

Section \ref{sec_versa} validates the adaptability of our method across diverse scenarios. This section will further demonstrate the adaptability of the proposed method when applied to different AUTs within the same scenario. Examples are as illustrated in Fig. \ref{exp-aut_adpt}. We maintain the same presentation style and organizational format as those used in Fig. \ref{exp-versa}.

\begin{figure}[pos=b] 
      \centering
      \includegraphics[width=.6\linewidth]{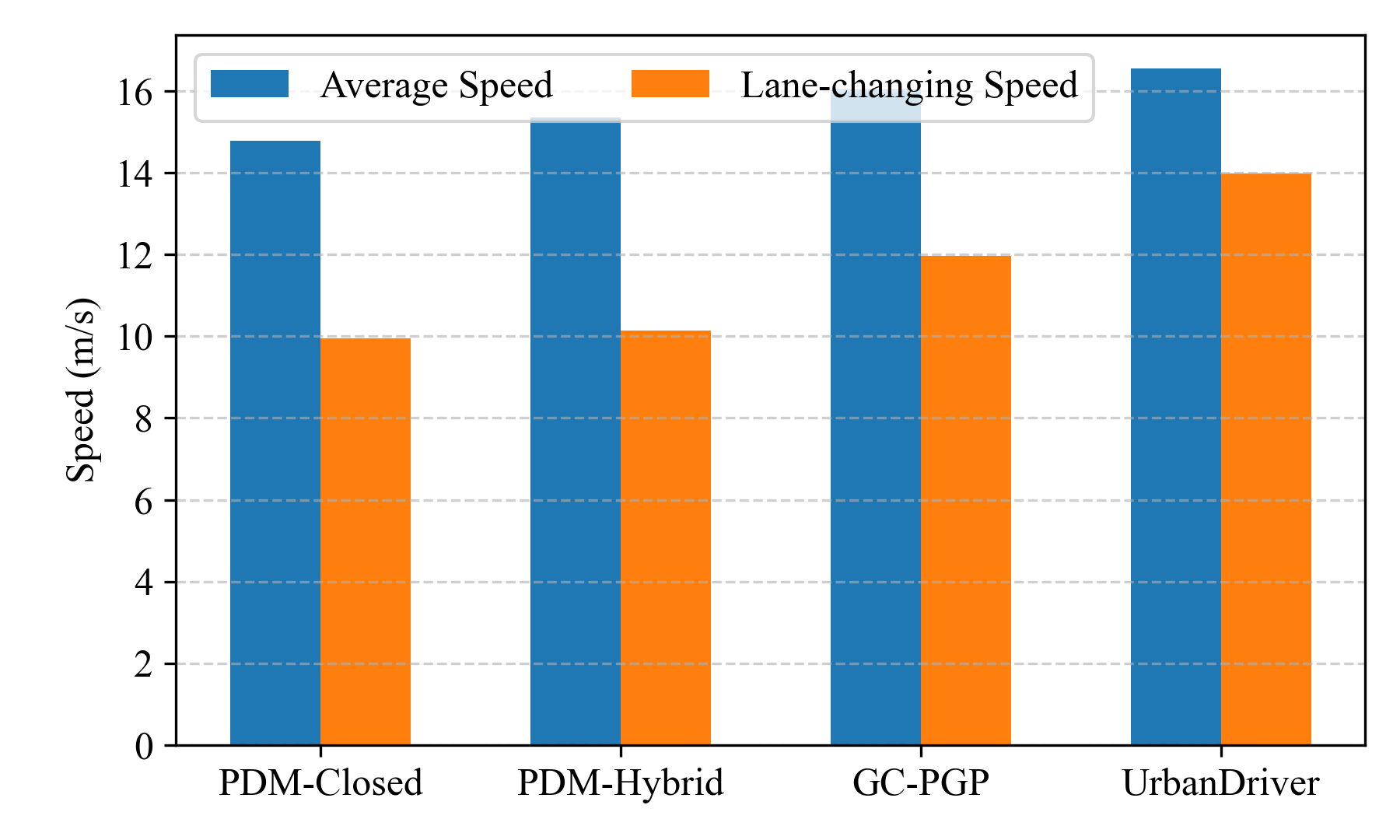}
      \caption{Comparative analysis of adversarial vehicle speed metrics across different AUTs.}
      \label{speed_comparison}
\end{figure}

As shown in Fig. \ref{exp-aut_adpt}, in the original scenario, both the SV and surrounding BVs maintain straight-driving within their respective lanes while approaching an intersection. Leveraging the pre-established accident knowledge database, the VLM selects BV No.1 in the adjacent lane ahead of the SV as the adversarial vehicle and guide it to execute a cut-in maneuver. The trajectory visualization demonstrates different adversarial trajectories generated by the guided diffusion model under the identical generation objective and guidance functions when applied to different AUTs. Specifically, the timing and aggressiveness of BV No.1's cut-in maneuvers vary significantly across different AUTs. When interacting with PDM-closed and PDM-Hybrid, BV No.1 is guided to execute earlier and more aggressive lane changes with faster lateral displacement. In contrast, when facing GC-PGP and UrbanDriver, the vehicle demonstrates delayed initiation and gentler lane changes with slower lateral movement.

To further quantify the performance of adaptive guidance, we analyze the average speed and lane-changing speed metrics of BV No.1 when interacting with different AUTs, with comparative results shown in Fig. \ref{speed_comparison}. As depicted in the figure, when the AUT is UrbanDriver, which tends to drive at higher speeds, the adaptive guidance framework proportionally increases both BV No.1's average speed and its instantaneous lane-change velocity. Conversely, when encountering more conservative AUTs like PDM-Closed, the system automatically adopts relatively lower cut-in speeds. The above analysis demonstrates that the proposed method can adaptively adjust to different AUTs, thereby enabling the generated safety-critical testing scenarios to effectively exert targeted testing pressure on algorithms with diverse behavioral patterns.

\subsection{VLM-directed generation validation}

\begin{figure}[pos=b] 
      \centering
      \includegraphics[width=\linewidth]{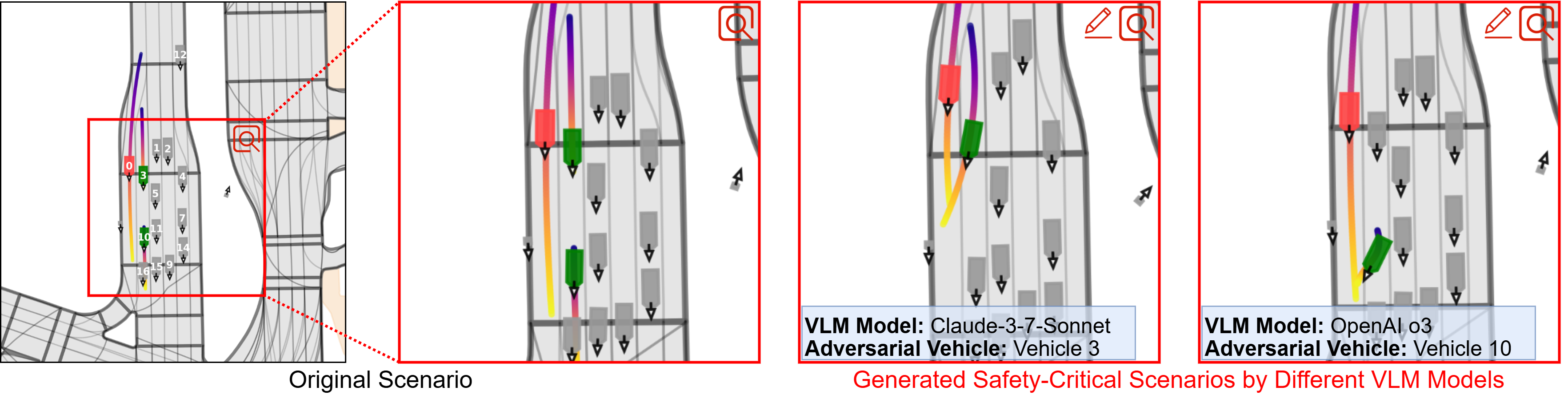}
      \caption{Examples of generating safety-critical scenarios using different VLMs under identical initial conditions and with the same AUT.}
      \label{exp-effect}
\end{figure}

As previously emphasized, to achieve VLM-directed safety-critical testing scenario generation, we grant the VLM full decision-making authority to serve as a strategist. Leveraging the prior accident knowledge database, it autonomously determines scenario generation objectives based on the current scenario state. As a result, employing different VLMs can yield varying understanding of scenarios, thereby producing diverse safety-critical testing cases. This section presents examples of generating safety-critical scenarios using different VLMs under identical initial conditions and with the same AUT, as illustrated in Fig. \ref{exp-effect}. In addition to the previously employed Claude-3-7-Sonnet, we also utilize OpenAI o3 as another VLM in this case study.

As shown in Fig. \ref{exp-effect}, in the original scenario, the SV travels in the rightmost lane approaching an intersection and prepares to make a right turn. BV No.3 moves in the straight-through lane to the left of the SV, gradually braking to a stop. BV No.10, also in the left straight-through lane relative to the SV, begins accelerating to proceed through the intersection, while other BVs maintain normal driving in other straight-through lanes. In the generated scenarios, Claude-3-7-Sonnet designates BV No.3 as the adversarial vehicle, guiding it to transition from gradual braking to continued movement followed by a right lane change. In contrast, OpenAI o3 selects BV No.10 as the adversarial vehicle, guiding it to initiate slow acceleration before making a sudden right turn when the SV approaches. Leveraging the prior accident knowledge database, both VLMs successfully identify potential cut-in risks present in the original scenario. The scenario generated by Claude-3-7-Sonnet demonstrates more natural and smooth vehicle dynamics, while OpenAI o3 produces a scenario with more abrupt maneuvers. Both scenarios remain plausible and demonstrate high testing value for ADS safety evaluation.

To summarize, following the concept of VLM-directed generation, the VLM is no longer just a user query translator, but a strategist that can fully utilize its reasoning capabilities. At the same time, our framework supports the integration of any VLM. With the rapid development and iteration of LM technology, more realistic, diverse, and highly interactive safety-critical testing scenarios are expected to be generated.

\subsection{Validation of VLM scenario understanding accuracy and ablation studies}

Throughout the scenario generation process, scenario understanding serves as one of the most fundamental and critical steps. It directly impacts the correctness and rationality of the subsequent guidance functions formulated by the VLM, and ultimately determines the quality of the generated safety-critical testing scenarios. To ensure accurate VLM comprehension, in Section \ref{sec_vlm} we have provided the VLM with multimodal inputs and implemented a CoT reasoning process. In this section, we will validate the accuracy of the VLM's scenario understanding and conduct ablation studies to verify the importance of each configuration component of the VLM.

To validate the accuracy of the VLM's scenario understanding, we first extract 30 diverse traffic scenario segments from the validation set of nuPlan as test samples, which are then meticulously annotated by human experts. The manual annotation assigns structured labels to each scenario, comprising three aspects: (1) SV behavior (e.g., going straight, turning left/right, lane changing); (2) road structure types (e.g., intersections, T-junctions, straight roads, ramps); and (3) key surrounding BVs' information (including vehicle ID lists, relative positions, and behavioral intentions). After that, these scenarios are fed into the VLM for scenario understanding across different configurations, and the VLM's comprehension accuracy is quantified. The experimental results are presented in Table \ref{tab5}. Note that for identification tasks with deterministic labels (SV behaviors, road structures and other elements), we directly compute accuracy using binary classification. For tasks involving ambiguous labels such as the relative positions and driving intentions recognition of surrounding BVs, we manually verify the consistency of results and calculate the accuracy using the formula: (number of correctly identified vehicles)/(total number of key surrounding BVs). 

As evidenced in Table \ref{tab5}, the full configuration demonstrates optimal performance across most evaluation dimensions, validating the efficiency of our multimodal input and CoT reasoning strategy. However, when identifying "Other elements" (e.g., pedestrians, crosswalks, traffic lights), the full configuration shows marginally inferior results compared to the non-CoT configuration. We hypothesize this minor discrepancy (while still maintaining high accuracy $\ge 90\%$) may stem from information conflicts caused by over-reasoning through CoT, as these elements are explicitly represented in both textual and visual inputs.

The non-BEV configuration exhibits notable performance degradation, particularly in the identification of "Road structure" (dropping from 91.7\% to 83.3\%) and "Surrounding BV intentions" (dropping from from 87.5\% to 66.7\%). This demonstrates that the global spatial-visual information provided by the BEV is crucial for the VLM to accurately understand road layouts and assess dynamic inter-vehicle relationships.

The non-textual description configuration exhibits a dramatic performance decline, demonstrating the poorest results across most evaluation dimensions, particularly in "Other elements" (merely 8.33\%) and "Surrounding BV intentions" (45.8\%). This result proves that structured textual descriptions serve as the primary carrier for conveying core semantic information of scenarios. While human experts can achieve efficient and accurate scenario understanding using only BEV images, current VLMs still show limitations in processing such visual data.

\begin{table}[width=\linewidth,cols=9,pos=b]
\renewcommand{\arraystretch}{1.3}
\caption{Experimental results on VLM scenario understanding accuracy and ablation studies.}
\label{tab5}
\begin{center}
\begin{tabular}{ m{0.03\linewidth}<{\centering} m{0.1\linewidth}<{\centering}  m{0.03\linewidth}<{\centering} m{0.005\linewidth} m{0.1\linewidth}<{\centering}  m{0.12\linewidth}<{\centering}  m{0.12\linewidth}<{\centering} m{0.12\linewidth}<{\centering} m{0.1\linewidth}<{\centering}}
\toprule
\multicolumn{3}{c}{VLM Configuration} & ~ & \multicolumn{5}{c}{Identification Accuracy} \\ \cline{1-3} \cline{5-9}
BEV & Textual descriptions & CoT & ~ & SV behavior & Road structures & Surrounding BV locations & Surrounding BV intentions & Other elements \\
\midrule
\usym{2713}  & \usym{2713} & \usym{2713} & ~ & \textbf{91.7\%} & \textbf{91.7\%} & \textbf{75.0\%} & \textbf{87.5\%} & 91.7\% \\
\usym{2717}  & \usym{2713} & \usym{2713} & ~ & 91.7 \% & 83.3\% & 75.0\% & 66.7\% & 66.7\% \\
\usym{2713}  & \usym{2717} & \usym{2713} & ~ & 50.0 \% & 75.0\% & 50.0\% & 45.8\% & 8.33\% \\
\usym{2713}  & \usym{2713} & \usym{2717} & ~ & 91.7 \% & 83.3\% & 33.3\% & 79.2\% & \textbf{95.8\%} \\

\bottomrule
\end{tabular}
\end{center}
\end{table}

The non-CoT configuration maintains relatively high accuracy for macro-level features like "SV behavior" (91.7\%) and "Road structure" (83.3\%). However, its performance sharply decreases to just 33.3\% on the "Surrounding BV locations" which requires more fine-grained identification at the micro-level. This significant drop demonstrates that without structured CoT guidance, VLMs struggle to effectively process inputs containing numerous vehicles and complex interaction patterns, making them prone to hallucinations and misjudgments about surrounding vehicles' states.

\section{Conclusion} \label{sec5}

This paper proposes a three-layer hierarchical scenario generation framework that transforms real-world driving data into safety-critical and highly interactive testing scenarios through guided diffusion, thereby ensuring both fidelity and diversity of the generated scenarios. The framework comprises a strategic layer driven by VLM for generation objective determination, a tactical layer that bridges high-level scenario generation objectives with underlying guidance functions, and an operational layer based on guided diffusion models for online generation and closed-loop simulation. 
Specifically, we first construct a high-quality fundamental diffusion model to learn the distribution of real-world scenarios. Then, we design an adaptive guided diffusion framework incorporating time domains and trigger conditions to ensure model adaptability. Finally, we integrate a VLM as the top-level strategist to perform scenario understanding and generation objective determination, thereby fully leveraging its powerful understanding and reasoning capabilities to achieve VLM-directed generation.

Experimental results on four distinct types of AUTs demonstrate that our proposed method significantly improves scenario safety-criticality, overall risk level, and task completion difficulty. Compared with original scenarios, the generated critical scenarios increase the average at-fault collision rate of the AUTs by approximately $4.2\times$. Furthermore, case studies validate the method's adaptability across diverse scenarios and different AUTs, as well as performance in VLM-directed generation when using various VLMs. Finally, ablation studies confirm the effectiveness of our multimodal input and CoT reasoning strategy for VLM utilization.

Since the types of generated safety-critical testing scenarios primarily depend on the prior accident knowledge database, future work will incorporate defensive driving scenarios \citep{NYDMV2020defensive} into the knowledge database to obtain more complex and diverse scenarios. Additionally, Retrieval-Augmented Generation (RAG) technology can be employed to upgrade VLMs into specialized safety-critical testing scenario generation experts. Furthermore, the proposed framework demonstrates the capability to guide multiple BVs and other traffic participants, enabling the generation of complex urban testing scenarios featuring multi-vehicle cooperative adversarial interactions in future implementations. Finally, with the rapid advancement of large model technologies, applying state-of-the-art LMs holds significant promise for substantially improving the method's VLM-directed generation performance.

\section*{Acknowledgment}
This study was supported by the National Key R\&D Program of China under Grant 2024YFB2505705, and the National Natural Science Foundation of China under Grant 52572482.

\printcredits

\bibliographystyle{cas-model2-names}

\bibliography{citelist}

\end{document}